\documentclass[preprint,12pt,authoryear]{elsarticle}

\usepackage{amssymb}
\usepackage{amsmath}
\setcitestyle{square}
\usepackage{booktabs}
\usepackage{float}
\usepackage{booktabs}
\usepackage{tabularx}
\usepackage{hyperref}
\usepackage[utf8]{inputenc}
\usepackage{array}
\usepackage{pdflscape} % Para rotar la página a horizontal
\usepackage{subcaption} % Para manejar la matriz de subfiguras

\begin{document}

\begin{frontmatter}



\title{Data-driven forecasting of tropical cloud forest phenology: A dynamical systems and machine learning hybrid approach} %% Article title


\author[a]{P. García} %% Author name
\affiliation[a]{organization={Departamento de Física, Facultad de Ingeniería, Universidad Católica Andrés Bello},
%Department and Organization
%            addressline={Urb. Montalban}, 
            city={Caracas},
%            postcode={1030}, 
%            state={Distrito Capital},
            country={Venezuela}}

\author[c]{S. Flores} %% Author name

\affiliation[c]{organization={Laboratorio de Ecología  de Suelos, Centro de Ecología, Instituto Venezolano de Investigaciones Científicas},
%Department and Organization
            city={Caracas},
            country={Venezuela}}

%\author[d]{H. Sulbaran} %% Author name
%\affiliation[d]{organization={ Doctorado en Modelamiento Matemático Aplicado, Universidad Católica del Maule},
%Department and Organization
%            city={Talca},
%            country={Chile}}

%% Abstract
\begin{abstract}
Forecasting how species and ecosystems respond to climate change is a critical challenge in ecology, particularly in hyper-diverse tropical biomes. While traditional phenological models often rely on preconceived functional forms, this study introduces a data-driven hybrid approach that integrates dynamical systems theory with machine learning to forecast flowering and fruiting cycles. 

Using a robust 35-year dataset from a Venezuelan tropical cloud forest, we reconstruct the system's state-space using delay vectors inspired by Takens' theorem, incorporating climate forcing variables such as temperature and precipitation. The model was trained using Gaussian Process Regression, providing both point predictors and explicit uncertainty estimates. 

Our results clearly suggest that the synergy between dynamical systems and machine learning effectively captures the non-linear complexities of ecological time series. Specifically, the model achieved higher predictive accuracy for flowering and fruiting. This hybrid framework offers a scalable tool for anticipating phenological mismatches and informing conservation strategies in the face of global environmental change.
\end{abstract}



%% Keywords

\end{frontmatter}




Phenological events, such as flowering and fruiting, are fundamental to ecosystem productivity and functioning. These cycles significantly impact biodiversity, food security, and human health by governing the timing of resource availability \cite{alberton, tscharntke, augustin}. In tropical cloud forests—environments characterized by persistent cloudiness, high humidity, and daily thermal fluctuations—phenology plays a synergistic role in the survival of endemic species \cite{williams1999,Bartholomew2026} and the maintenance of complex trophic interactions \cite{sullivan2023cascading}. However, contemporary environmental changes are triggering phenological mismatches: shifts in timing that decouple plants from their pollinators or dispersers, potentially altering nutrient cycling and community structures \cite{fu2022plant, bucher2021timing}. Forecasting these shifts is therefore essential to anticipate ecological disruptions and develop robust conservation strategies.

The challenge of forecasting in these "super-speciose biomes" lies in the inherent complexity of their ecological processes. Tropical cloud forests can be conceptualized as dynamical systems: complex entities where internal biological mechanisms and external climatic drivers interact through non-linear feedback loops \citep{li2023machine}. In such systems, minor variations in climatic input can result in significant, often unpredictable, shifts in phenological output. To address this, dynamical systems theory provides a rigorous mathematical framework—often utilizing differential or difference equations—to describe how the state of a system evolves over time \cite{Garcia-2022,Garcia-1996}. By adopting a State-Space Ecological Model approach, we can represent the current state of the forest through a set of variables that change dynamically, allowing us to simulate ecosystem behavior under varying environmental scenarios \citep{auger2021guide}.

While dynamical systems offer the structural framework, the high dimensionality and non-linearity of ecological data often exceed the limits of traditional mechanistic models. This is where Machine Learning (ML) becomes a powerful complement. ML strategies enable the modeling of complex functions without requiring a preconceived functional form \citep{weinan2017proposal}, facilitating the analysis of noisy or incomplete time-series data through discretization and optimization \citep{pourmohammad2023dynamical}. However, applying ML to tropical phenology is not without hurdles. The accuracy of predictions is frequently hampered by data gaps, the non-linear influence of precipitation and light intensity \citep{guan2023study}, and the difficulty of ensuring that models trained on historical data can generalize to unprecedented future climatic conditions \citep{Wang2019}.

To mitigate these challenges, advanced techniques such as Gaussian Process Regression (GPR) have emerged as vital tools \cite{Ramusen}. Unlike "black-box" algorithms, GPR provides explicit uncertainty estimates, such as confidence intervals, which are crucial for evaluating the reliability of ecological forecasts \citep{verrelst2013gaussian}. When combined with rigorous validation methods like cross-validation \citep{yates2023cross}, these models allow for a more nuanced interpretation of how tropical forests respond to environmental change.

In this paper, we propose a data-driven phenological model that bridges the gap between dynamical systems theory and machine learning. Our approach avoids assuming preconceived functional dependencies, instead allowing the underlying patterns to emerge from the interaction between climatic and phenological data. The manuscript is organized as follows: first, we describe the study site and the 35-year dataset \citep{flores2023extreme} of flowering and fruiting cycles. Second, we detail the mathematical architecture of our state-space model and the ML strategies used for its training. Finally, we present a detailed analysis of our results, followed by a discussion on the implications for tropical forest conservation.

\section{Material and methods}
\subsection{The dynamical system: a tropical cloud forest}
The research site, known as Altos de Pipe ($10.397$, $-66.985$), is located along the northern ocean-facing slopes of the {\it Cordillera de la Costa} northern Venezuela. This site, spanning an area of approximately $1~ km^2$, is perched at an elevation of $1747~ m$ above sea level and is a mere $11~ km$ from Caracas. The region is characterized by a mean annual precipitation of $1200~ mm$, with a clear demarcation of dry (January to March) and rainy (June to November) seasons. Transitional periods are observed from wet-to-dry during November–December and dry-to-wet during April–May. The site maintains a mean annual temperature of $19~ ^oC$.

The study site is typified as a cloud forest, composed of three distinct strata. The emergent stratum, extending to a height of $25--30 m$, is dominated by species such as {\it Aspidosperma fendleri} Woodson (Apocynaceae) and {\it Podocarpus salicifolius} Klotzsch \& H.Karst. ex Endl (Podocarpaceae). The canopy, or the second stratum, stands at approximately $15~ m$ in height and is home to a diverse mix of species, including {\it Richeria grandis}, Vahl (Phyllanthaceae) and {\it Protium tovarense}, Pittier (Burseraceae). The understory stratum is densely populated with shrubs from the Melastomataceae and Rubiaceae families, including the melastome shrub, {\it Miconia dodecandra} Cogn, and the rubiaceous shrub, {\it Palicourea fendleri} Standl. For a more comprehensive understanding of the species studied and the characteristics of the study site, refer to \cite{flores2023extreme}.

\subsection{The data from the system}
Phenology and climate For $35$ continuous years, every month, the reproductive status of five individuals in $15$ focal tree species (Table \ref{species}) in 12 families was recorded; distances between individuals within a species were $5--20 m$ and the trees were tagged and numbered for monthly observations and subsequent identification, and voucher specimens were deposited at the herbarium of the Instituto Venezolano de Investigaciones Científicas (IVIC). Climate data were collected from a station  (INAME IVIC 82-90) within 1 km of the study site, managed by the Centro de Ciencias Atmosfericas y Biogeoquímica (reconstructed by \cite{kalnay2018ncep}); these data were available as monthly values for temperature (mean average degree Celsius temperature per month), precipitation (mm/mo), and soil moisture (volumetric percentage). 

%For additional information regarding the plant species' characteristics and the environmental data related to the cloud forest study site, please refer to (Appendix: Tables S1 and S4 in \citep{flores2023extreme}).

\begin{table}[H]
\centering
\vspace{2mm}
\setlength{\tabcolsep}{3pt} % Reduce el espacio entre columnas
\scriptsize % Un poco más grande que tiny para que sea legible
\begin{tabularx}{\textwidth}{c l >{\itshape}X c c c}
\toprule
\textbf{No.} & \textbf{Family} & \textbf{Species} & \textbf{GBIF link} & \textbf{Growth form} & \textbf{Reproductive density} \\
 & & & & & \textbf{\tiny(4m$^{-2}$)} \\
\midrule
1  & Apocynaceae      & Aspidosperma fendleri Woodson               & \href{https://www.gbif.org/species/3618540}{3618540} & Tree    & 0.25 $\pm$ 0.18 \\
2  & Burseraceae      & Protium tovarense Pittier                  & \href{https://www.gbif.org/species/3992944}{3992944} & Tree    & 0.29 $\pm$ 0.21 \\
3  & Phyllanthaceae   & Hieronyma fendleri Briq.                   & \href{https://www.gbif.org/species/3079076}{3079076} & Tree    & 0.20 $\pm$ 0.16 \\
4  & Phyllanthaceae   & Richeria grandis Vahl                      & \href{https://www.gbif.org/species/3079144}{3079144} & Tree    & 0.23 $\pm$ 0.27 \\
5  & Marcgraviaceae   & Ruyschia tremadena (Ernst) Lundell         & \href{https://www.gbif.org/species/3915996}{3915996} & Climber & 0.10 $\pm$ 0.01 \\
6  & Malpighiaceae    & Byrsonima karstenii W.R.Anderson           & \href{https://www.gbif.org/species/3985626}{3985626} & Tree    & 0.12 $\pm$ 0.08 \\
7  & Hypericaceae     & Vismia ferruginea Kunth                    & \href{https://www.gbif.org/species/7679156}{7679156} & Tree    & 0.29 $\pm$ 0.23 \\
8  & Rubiaceae        & Palicourea fendleri Standl.                & \href{https://www.gbif.org/species/2899131}{2899131} & Shrub   & 0.66 $\pm$ 0.88 \\
9  & Nyctaginaceae    & Guapira olfersiana (Link, et al.) Lundell & \href{https://www.gbif.org/species/3701055}{3701055} & Tree    & 1.38 $\pm$ 0.75 \\
10 & Erythroxylaceae  & Erythroxylum amazonicum Peyr.              & \href{https://www.gbif.org/species/3694127}{3694127} & Shrub   & 3.20 $\pm$ 1.12 \\
11 & Melastomataceae  & Miconia dodecandra (Desr.) Cogn.           & \href{https://www.gbif.org/species/3867434}{3867434} & Tree    & 0.21 $\pm$ 0.19 \\
12 & Melastomataceae  & Graffenrieda latifolia (Naudin) Triana     & \href{https://www.gbif.org/species/5600001}{5600001} & Tree    & 0.52 $\pm$ 0.26 \\
13 & Phyllanthaceae   & Tetrorchidium rubrivenium Poepp.           & \href{https://www.gbif.org/species/3062925}{3062925} & Tree    & 0.60 $\pm$ 0.31 \\
14 & Malvaceae        & Heliocarpus americanus L.                  & \href{https://www.gbif.org/species/4259460}{4259460} & Tree    & 0.16 $\pm$ 0.10 \\
15 & Adoxaceae        & Viburnum tinoides var. venezuelense...     & \href{https://www.gbif.org/species/7881568}{7881568} & Tree    & 0.23 $\pm$ 0.15 \\
\bottomrule
\end{tabularx}
\caption{Focal study species.}
\label{species}
\end{table}

\subsection{The model}
Let us assume the forest section as a dynamical system, that is, a set $\{S,F,T\}$ where $S$ is the set of states accessed by the system or in other words, information available about the system, $F$ is the dynamical rule governing the evolution of the states and $T$, the time, is a set of integer numbers in this case.
	
From that perspective, if we assume that the data comes from monitoring a dynamical system given by the $"forest = species~ +~ climate"$, it is necessary to specify what we will understand as state.
Thus, we will suppose that the state will be given by the concatenation of the climatological state variables ${\bf x}_t$ and the phenological variables $({\bf y}^j_t, {\bf z}^j_t)$ corresponding to the number of flowers and fruits respectively, of the specie $i$ at time $t$ (considered discrete in this case). Thus, we will define the vectors,
\begin{eqnarray}
	{\bf x}^i_t &=& ( x^i_{t}, x^i_{t-1}, x^i_{t-2}, \cdots, x^i_{t-m_{1,i}} ), ~~~~i= 1, 2, \cdots, m,\nonumber \\
	{\bf y}^j_t &=& ( y^j_{t}, y^j_{t-1}, y^j_{t-2}, \cdots, y^j_{t-m_{2,j}} ), ~~~~j= 1, 2, \cdots, k, \nonumber \\
	{\bf z}^j_t &=& ( z^j_{t}, z^j_{t-1}, z^j_{t-2}, \cdots, z^j_{t-m_{3,j}} ), ~~~~j= 1, 2, \cdots, k.
\end{eqnarray}

\noindent
where $m$ is the number of climatological variables to be considered, $k$ is the number of species to be monitored and  $m_{i,j}$, $i=1,2,3$ are the dimensions of the delay vectors of the climatological variables and each monitored specie.

The delay, in vectors representing the phenological or climatological states, are inspired by those used in the reconstruction of attractors of chaotic systems, proposed by Takens\cite{takens2006detecting} and that has been useful in the characterization of biological systems, such as cardiac or neurological dynamics, among others.

The dynamical rule in this case, can be thought of as:
\begin{eqnarray}
	x^i_{t+1} &=& f_i({\bf x}_t) + \epsilon~ u({\bf y}^j_t, {\bf z}^j_t), \nonumber \\
	y^j_{t+1} &=& g_j({\bf x}_t, {\bf y}^j_t),  \nonumber \\
	z^j_{t+1} &=& h_j({\bf x}_t, {\bf y}^j_t, {\bf z}^j_t),
	\label{Dinamical System}
\end{eqnarray}

\noindent
where, $f_i$, $g_j$ and $h_j$ represent the dynamic rules governing the $i$-th climate variable,  flowering and fruiting of the $j$-th species, respectively. And $u$ expresses the effect of phenology on climate, so we select $epsilon \to 0$, so that the perturbation to climate by phenology is very small. 
Note how, the last two equations in (\ref{Dinamical System}), reflect the causal relationship between flowering and fruiting, at least in the short term. 

This modeling strategy is frequently used for design models of biological systems and is referred to as state-space modeling of ecological time series, see \cite{davis2022new} and references therein.

Finally, the task of predicting system evolution consists of finding approximations for the non-autonomous and parameterized column vector valuated dynamic ${\bf F} = ( F_1, F_2, \cdots, F_k)^T$, corresponding to each of the $k$ monitored species:

\begin{equation}
{\bf s}^{j}_{t+1} = F_j({\bf s}^{j}_t),~~~ j=1,2,\cdots, k
\label{Gobal-dim}
\end{equation}

\noindent
where the dynamical rule $F_j$ is given by:  

\begin{equation}
F_j({\bf s}^j_t) = (g_j({\bf x}_t, {\bf y}^j_t), h_j({\bf x}_t, {\bf y}^j_t, {\bf z}^j_t))^T,
\end{equation}

\noindent
given the phenological  history ${\bf s}^{j}_t = ({\bf y}^j_t, {\bf z}^j_t)$ and a the climate forcing climatological variable ${\bf x}_t$.

\subsection{Fitting the model}

The challenge of forecasting tropical phenology can be formally structured as a nonlinear regression problem. Following the dynamical systems framework, our goal is to estimate the functions $g$ and $h$ that map the current reconstructed state ${\bf s}^{j}_{t}$ to its subsequent evolution ${\bf s}^{j}_{t+1}$. Estimating these relationships is typically accomplished by constraining a class of test functions and then optimizing the function parameters by using training data to produce the smallest error, a strategy previously explored in phenological dynamics by \cite{Zalika}.

\subsubsection{From Kernel Ridge Regression to Gaussian Processes Regression}

To handle the inherent non-linearity of ecological data, we transition from simple linear models to a functional space of higher dimension. In a deterministic setting, this is addressed via Kernel Ridge Regression (KRR) \cite{cristianini2000introduction}. The fundamental idea is to map the input data $x$ into a high-dimensional (potentially infinite) feature space $\mathcal{F}$ through a mapping function $\phi(x)$. In this augmented space, we seek a predictor $f$ that can be expressed as a linear combination of the basis elements of the Reproducing Kernel Hilbert Space  $\mathcal{H}$:
\begin{equation}
f(x) = \langle w, \phi(x) \rangle_{\mathcal{F}} = \sum_{j} w_j \phi_j(x)
\end{equation}

Given a dataset $\mathcal{D} = \{(x_i, y_i)\}_{i=1}^N$, KRR seeks to find the function $f \in \mathcal{H}$ that minimizes the regularized loss:
\begin{equation}
\min_{f \in \mathcal{H}} \sum_{i=1}^N (y_i - f(x_i))^2 + \lambda \|f\|_{\mathcal{H}}^2
\end{equation}

According to the \textit{Representer Theorem} \cite{kimeldorf1971some, scholkopf2001generalized}, the solution $w$ that minimizes this cost function lies in the span of the training data points mapped into the feature space. That is, $w$ can be written as a linear combination: $w = \sum_{i=1}^N \alpha_i \phi(x_i)$. By substituting this expression into the definition of $f(x)$, we obtain:
\begin{equation}
f(x^*) = \sum_{i=1}^N \alpha_i \langle \phi(x_i), \phi(x^*) \rangle_{\mathcal{F}} = \sum_{i=1}^N \alpha_i K(x_i, x^*)
\end{equation}

This derivation introduces the \textit{``kernel trick''}, which allows us to compute complex non-linear relationships using only the inner product $K(x, x') = \langle \phi(x), \phi(x') \rangle_{\mathcal{F}}$ without ever explicitly defining the mapping $\phi$ \cite{Garcia-2007}. Thus, the function is represented as a weighted sum of the kernel evaluated at the training points.

However, while KRR provides a point estimate, it lacks a formal framework for uncertainty. This is where the Bayesian formalism of Gaussian Processes (GP) becomes fundamental \cite{Williams1998}. A GP generalizes KRR by treating the function $f(x)$ as a stochastic process, where any finite collection of values follows a multivariate Gaussian distribution.

\subsubsection{Bayesian Formalism and Predictive Inference}

Following \cite{Ramusen}, we define a prior distribution over functions,

\begin{equation}
f \sim \mathcal{GP}(m(x), k(x, x')).
\end{equation}

Assuming a zero-mean prior for simplicity, the joint distribution of the observed outputs $\mathbf{y}$ (with additive noise $\epsilon \sim \mathcal{N}(0, \sigma_n^2)$) and the function value $f_*$ at a test point $x^*$ is given by:
\begin{equation}
\begin{bmatrix} \mathbf{y} \\ f_* \end{bmatrix} \sim \mathcal{N}\left(\mathbf{0}, \begin{bmatrix} \mathbf{K} + \sigma_n^2\mathbf{I} & \mathbf{k}_* \\ \mathbf{k}_*^\top & k(x^*, x^*) \end{bmatrix}\right)
\end{equation}
where $\mathbf{K}$ is the Gram matrix ($K_{ij} = k(x_i, x_j)$) and $\mathbf{k}_*$ is the vector of covariances between the test point and training points.

The relationship between KRR and GP is revealed in the predictive posterior distribution $P(f_* | \mathbf{X}, \mathbf{y}, x^*)$. By applying the conditioning rules for Gaussians, the posterior mean is:
\begin{equation}
\bar{f}_* = \mathbf{k}_*^\top (\mathbf{K} + \sigma_n^2\mathbf{I})^{-1} \mathbf{y}
\end{equation}
Note that if we set the regularization parameter $\lambda = \sigma_n^2$, the GP posterior mean is mathematically identical to the KRR solution. The crucial advantage of the GP formalism lies in the posterior covariance:
\begin{equation}
\text{cov}(f_*) = k(x^*, x^*) - \mathbf{k}_*^\top (\mathbf{K} + \sigma_n^2\mathbf{I})^{-1} \mathbf{k}_*
\end{equation}

This expression provides an explicit measure of the model's confidence. In regions of the state space where data is sparse, the predictive variance increases, signaling higher uncertainty. This Bayesian integration of the kernel trick allows our hybrid model not only to track the non-linear dynamics of flowering and fruiting but also to quantify the reliability of the forecast, a feature that deterministic KRR cannot provide, as discussed in the context of tropical cloud forests by \cite{williams1999}.

For these Regression models, the choice of kernel function is critical, as it defines the expected smoothness and correlation properties of the function we are trying to learn. 
In this work, a combined kernel was chosen as the kernel function, defined as a constant kernel multiplied by an RBF kernel, plus a white kernel. This structure allows us to model the overall trend, the smooth long-term relationships, and the noise in the time series. The general form is: 

\begin{equation*} 
k_{\text{Combined}}(\mathbf{x}_i, \mathbf{x}_j) = k_{\text{Constant}}(\mathbf{x}_i, \mathbf{x}_j) \cdot k_{\text{RBF}}(\mathbf{x}_i, \mathbf{x}_j) + k_{\text{White}}(\mathbf{x}_i, \mathbf{x}_j) 
\end{equation*}

Here, $$k_{\text{RBF}}(\mathbf{x}_i, \mathbf{x}_j) = \sigma^2 \exp\left(-\frac{\|\mathbf{x}_i - \mathbf{x}_j\|^2}{2l^2}\right),$$

\noindent
and  also: $k_{\text{White}}(\mathbf{x}_i, \mathbf{x}_j) = \sigma_n^2 \delta_ {ij} $ and  $k_{\text{Constant}}(\mathbf{x}_i, \mathbf{x}_j) = c^2 $.

This selection is justified as a common and robust approach for time series that exhibit noise and, potentially, an underlying trend or mean. Specifically, the RBF Kernel is suitable for modeling smooth, long-term relationships; the White Kernel captures the noise inherent in the data; and the Constant Kernel allows for adjusting the overall mean of the function. Together, these components offer the flexibility to capture the complex dynamics of phenological series.

\section{Results}
In this section, we evaluate the predictive performance of the Gaussian Process Regression (GPR) model in forecasting both mean and species-specific phenological cycles. 
Following the optimization process during training, the kernel parameters obtained for each model were: (i) for the flowering model, $\sigma = 0.348$, $l = 0.571$, and $\sigma_n = 0.0146$; and (ii) for the fruiting model, $\sigma = 0.748$, $l = 1.96$, and $\sigma_n = 0.00836$.
These optimized parameters are the result of the process of fitting the model to the data, seeking the kernel configuration that best explains the variability and structure of the phenological series for flowering and fruiting.

The results are organized as follows: we first present the forecasting of mean forest phenology to capture the aggregate ecosystem response, followed by a detailed assessment of individual species. This dual analysis allows us to examine how the integration of dynamical systems and machine learning handles the varying degrees of periodicity and stochasticity inherent in tropical flowering and fruiting data.

\subsection{Data pre-processing}
In our case, the data are pre-processed, taking a minimum {\it moving-average} of $2$ elements to foresee that abrupt jumps in the data, in this case the presence of many zeros, will seriously affect the quality of the predictions. Subsequently the data are standardized, it mean, to convert the data to a time series with mean zero and standard deviation one.

Later, in order to evaluate its performance on unseen data later on, the data is divided into two subsets: the first one with  eighty percent of the total, used to optimize the model parameters, and the second one with twenty percent of the total data, used to validate the resulting model. 

\subsection{Forecasting mean phenology}
By “average phenology,” we mean that we will consider the arithmetic mean of the number of flowers and fruits across all monitored species and the average of all climate variables as the state of the system.\\

\begin{eqnarray}
	\overline x^{j}_t &=& \frac{1}{l}\sum_{j=1}^{l}{x^j_{t}}, \nonumber \\
	\overline y^{j}_t &=& \frac{1}{k}\sum_{j=1}^{k}{y^j_{t}}, \nonumber \\
	\overline z^{j}_t &=& \frac{1}{k}\sum_{j=1}^{k}{z^j_{t}},
\end{eqnarray}

\begin{figure}[h!]
	\begin{center}
		\includegraphics[width=14cm]{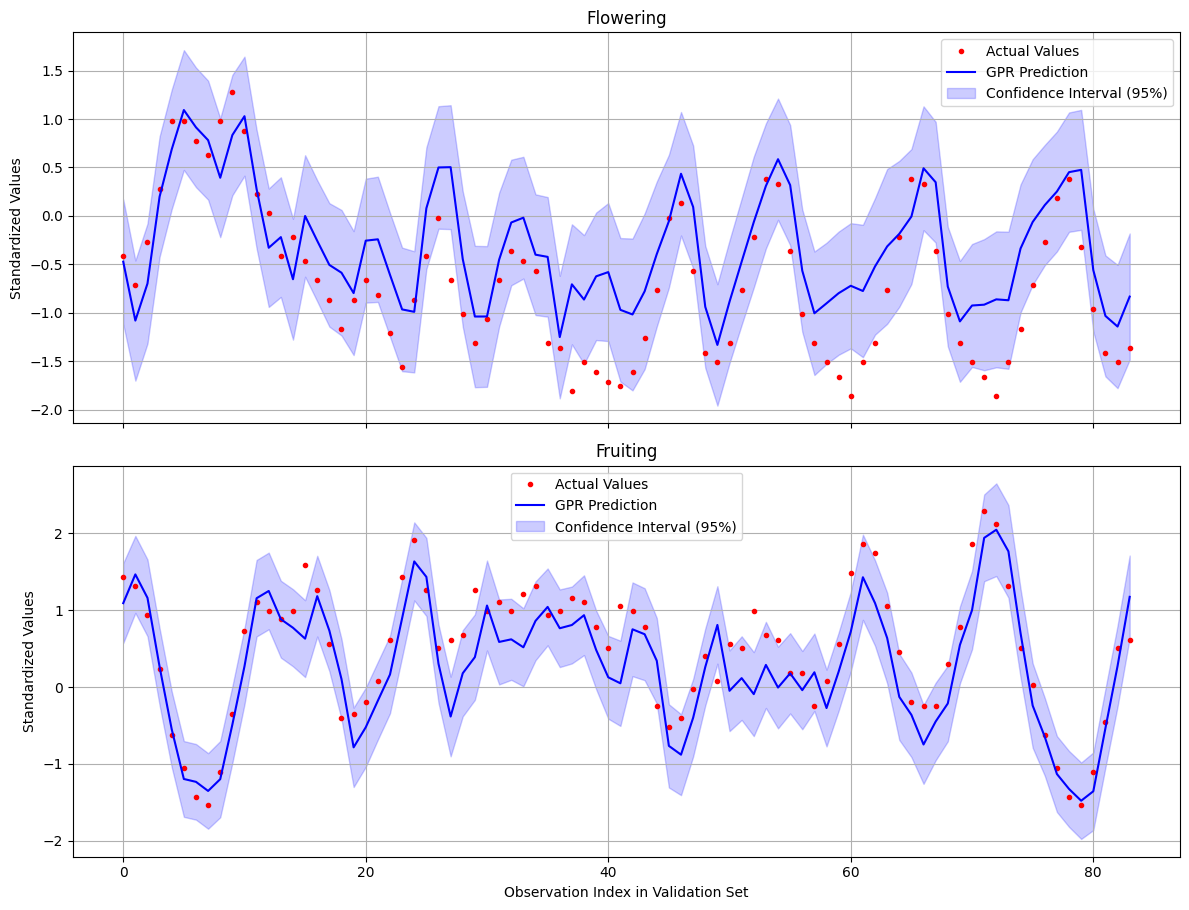}
		\caption{Forecasting mean phenology.}
		\label{MeanPred}
	\end{center}
\end{figure}

In this case, $k$ is the number of species being monitored, and $l$ is the number of climate variables being monitored. Figure \ref{MeanPred} shows the \textit{trajectory} of the dynamic system, where the states of the system’s components have been averaged and the data have been standardized so that the climate, flowering, and fruiting data are on similar scales.

\begin{figure}[h]
	\begin{center}
		\includegraphics[width=10cm]{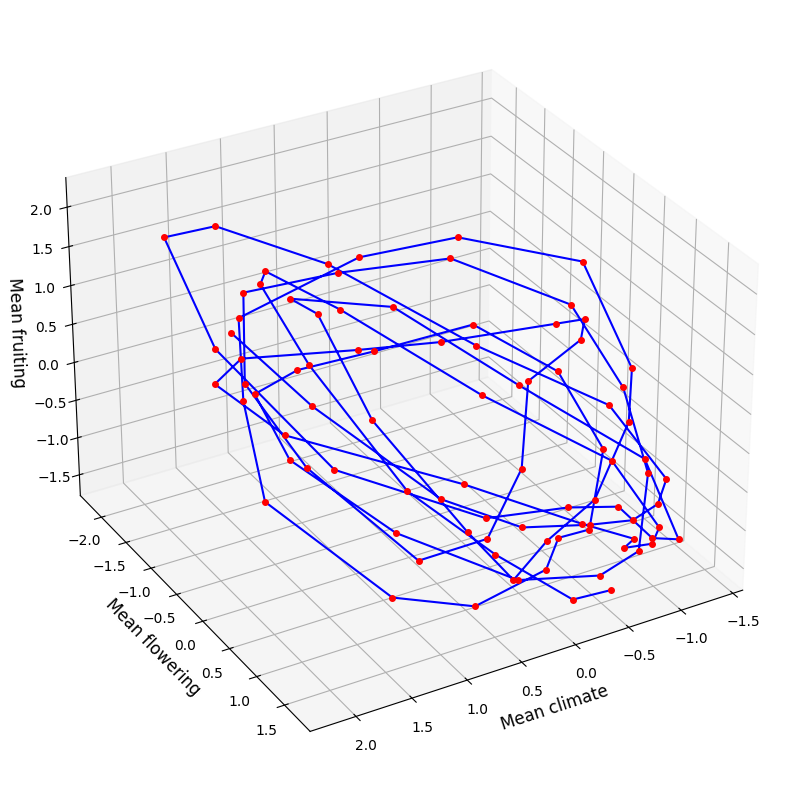}
		\caption{State space of the system forest-clima.}
		\label{Figx}
	\end{center}
\end{figure}

The mild evolution of the dynamic system, defined as the combination of forest phenology and climate and represented by flowering, fruiting, and climate variables—suggests that at least partially deterministic models can be used to predict the system’s evolution.

\subsection{Forecasting phenology of individual species}

Since we have defined our dynamic system as (flowering-fruiting-climate), to be consistent with Takens’ results and to account for the other variables that could affect flowering and fruiting but were not measured, we implement a state space reconstruction using delay vectors,  in such a way that 

\begin{eqnarray}
{\bf x}^j_t &=& ( x^j_{t}, x^j_{t-1}, x^j_{t-2}, \cdots, x^j_{t-m_{1,j}} ) \nonumber \\
{\bf y}^j_t &=& ( y^j_{t}, y^j_{t-1}, y^j_{t-2}, \cdots, y^j_{t-m_{2,j}} ) \nonumber \\
{\bf z}^j_t &=& ( z^j_{t}, z^j_{t-1}, z^j_{t-2}, \cdots, z^j_{t-m_{3,j}} ), 
\end{eqnarray}

In this case, the vectors representing the states of two dynamic systems (flowering and fruiting) are constructed by concatenating the climate lag vectors with a flowering lag vector (${\bf x}_t \oplus {\bf y}^j_t$) in the case of the flowering model, and in the case of fruiting, by concatenating the climate, flowering, and fruiting delay vectors (${\bf x}_t \oplus {\bf y}^j_t \oplus {\bf z}^j_t$). Thus, the dynamics will be given by

\begin{eqnarray}
	y^j_{t+1} &=& g({\bf x}_t, {\bf y}^j_t),  \nonumber \\
	z^j_{t+1} &=& h({\bf x}_t, {\bf y}^j_t, {\bf z}^j_t),
	\label{Mean Dinamical System}
\end{eqnarray}

Where we consider the climate as an autonomous dynamic that perturbs the dynamics of the forest and where  $m_{i,j}$, $i=1,2, 3$, are the delays corresponding to climate, flowering and fruiting respectively.\\

Finally, regarding the embedding dimension $m_{1,j}$, we will choose $m_{1,j}=m_{2,j}=m_{3,j}=3$, although we believe, based on the experiments we have conducted, that choosing a different embedding dimension for each species and for the climate could improve the quality of the predictions; in this case, the aim is simply to show that even when this parameter is not optimized, the results suggest that the hybrid approach combining dynamic systems and machine learning yields promising results.

\begin{figure}[ht!]
    \centering

    % FILA 1
    \begin{subfigure}{0.31\linewidth}
        \includegraphics[width=\textwidth]{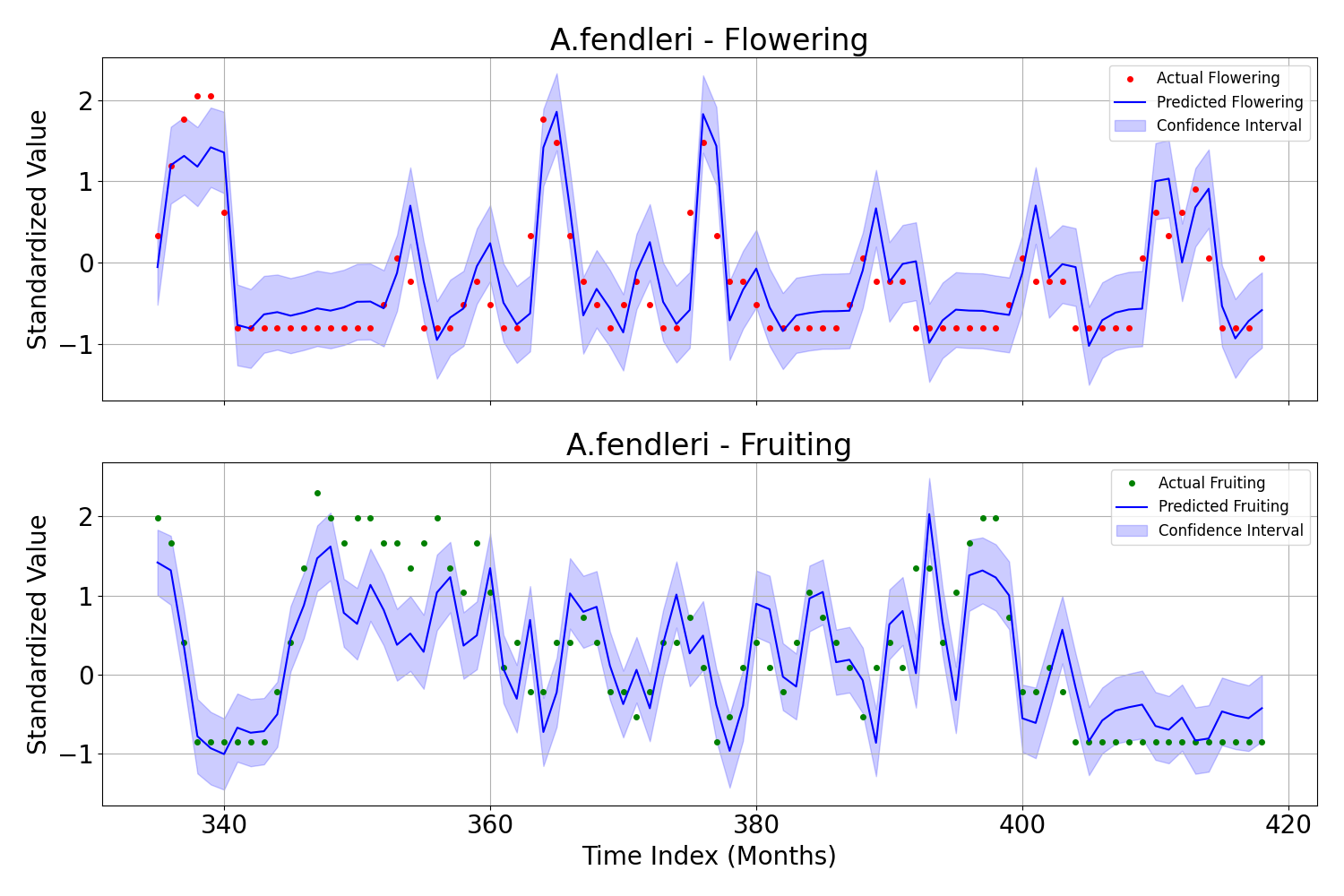}
    \end{subfigure}\hfill
    \begin{subfigure}{0.31\linewidth}
        \includegraphics[width=\textwidth]{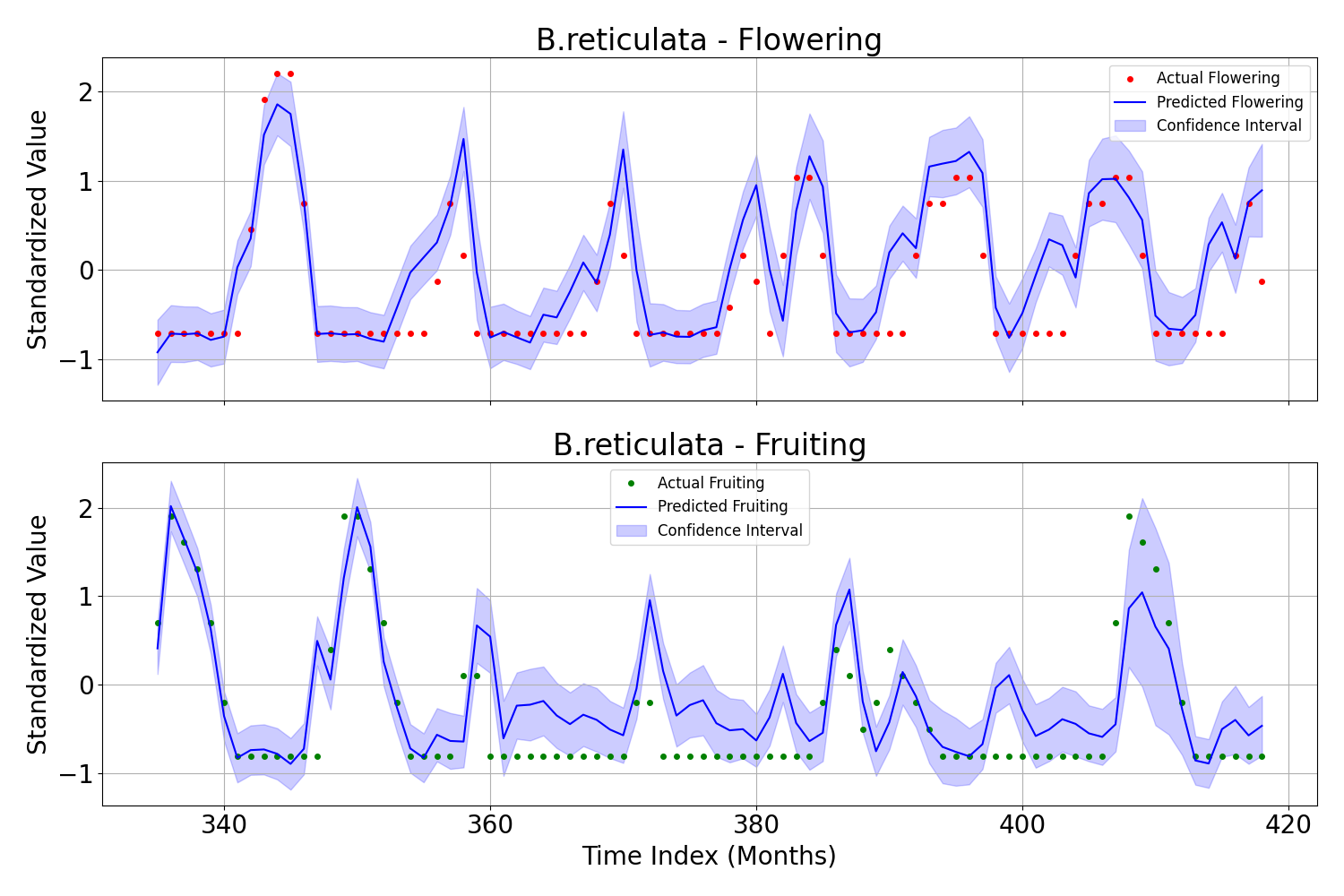}
    \end{subfigure}\hfill
    \begin{subfigure}{0.31\linewidth}
        \includegraphics[width=\textwidth]{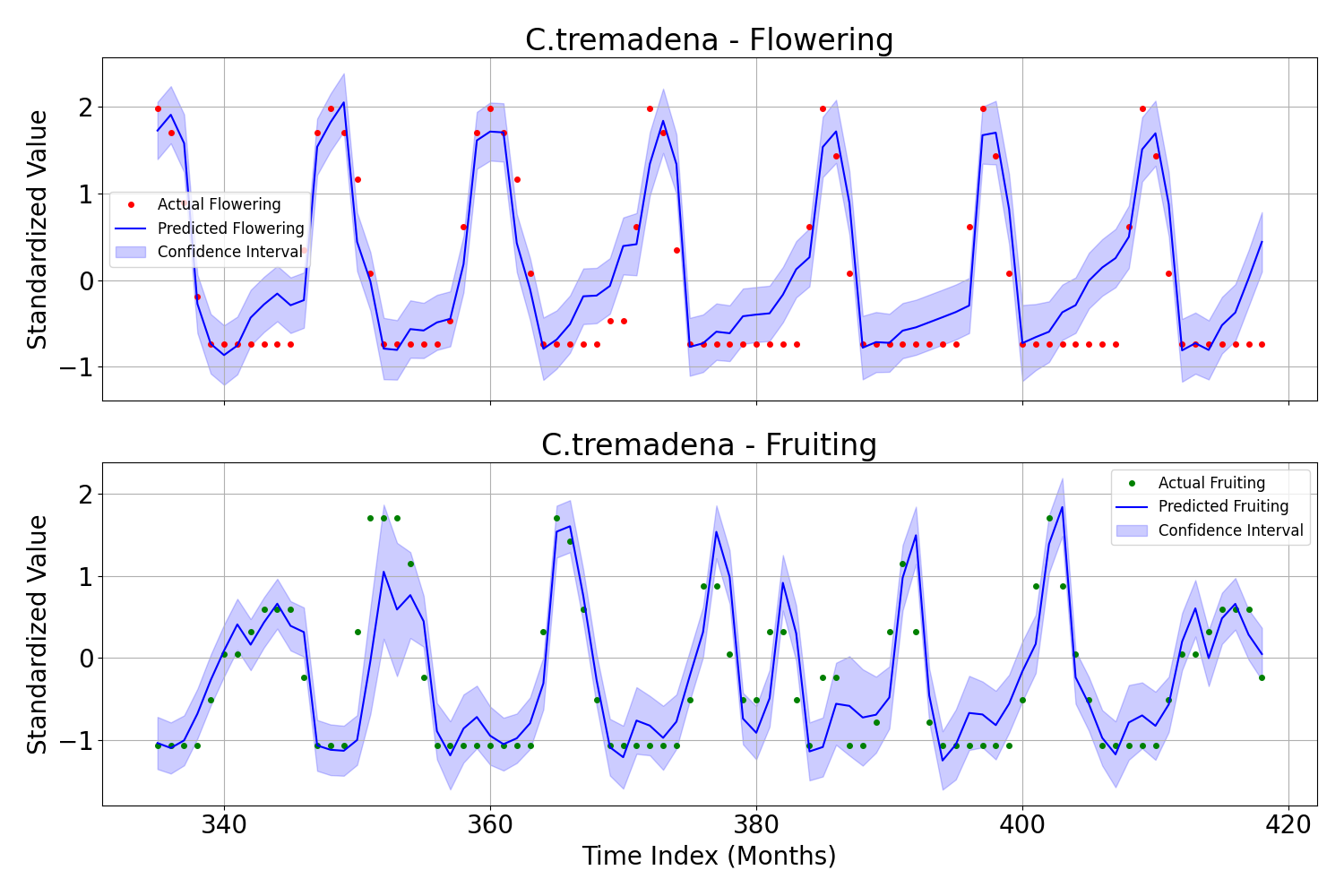}
    \end{subfigure}

    \vspace{0.2cm}

    % FILA 2
    \begin{subfigure}{0.31\linewidth}
       \includegraphics[width=\textwidth]{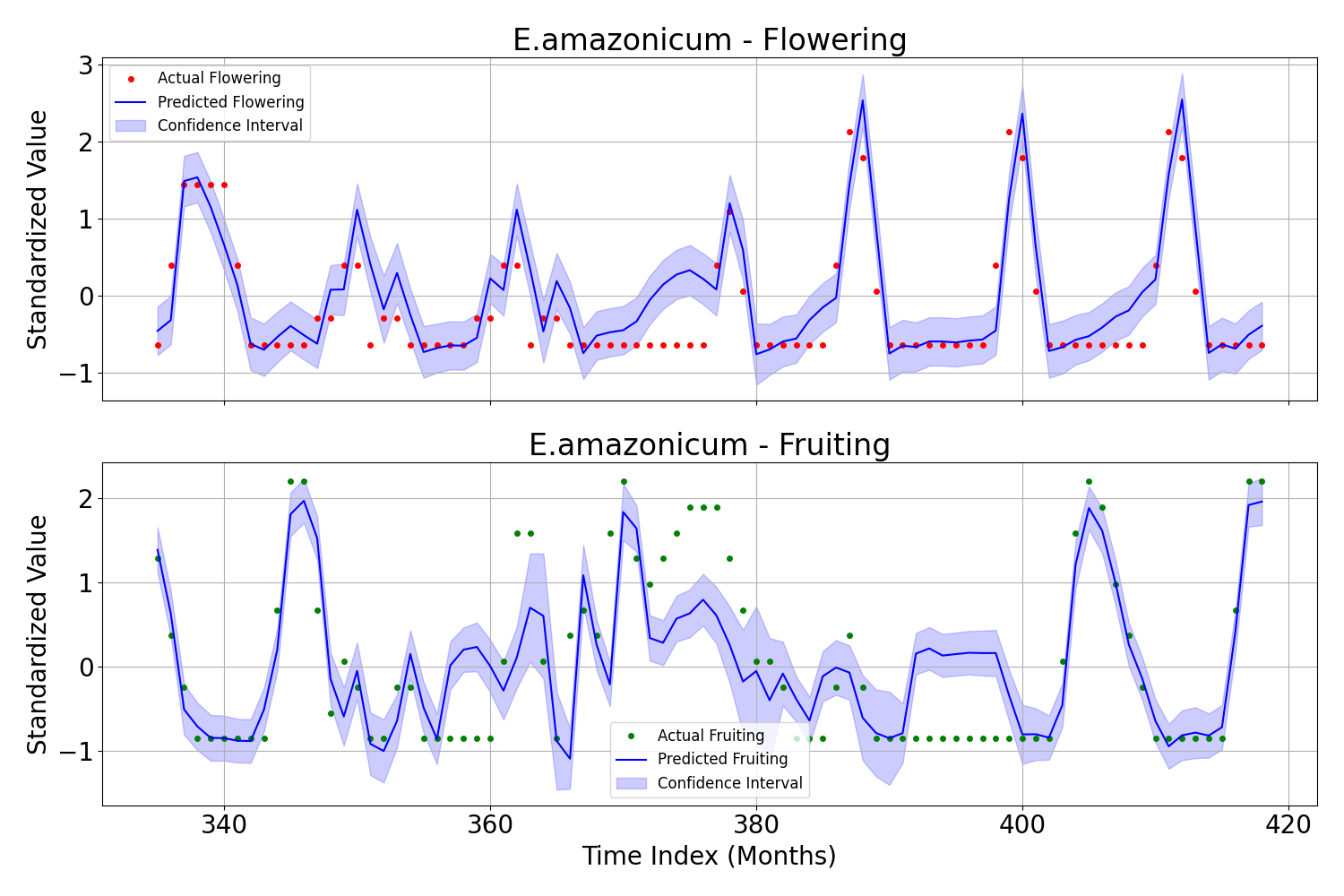}
    \end{subfigure}\hfill
    \begin{subfigure}{0.31\linewidth}
        \includegraphics[width=\textwidth]{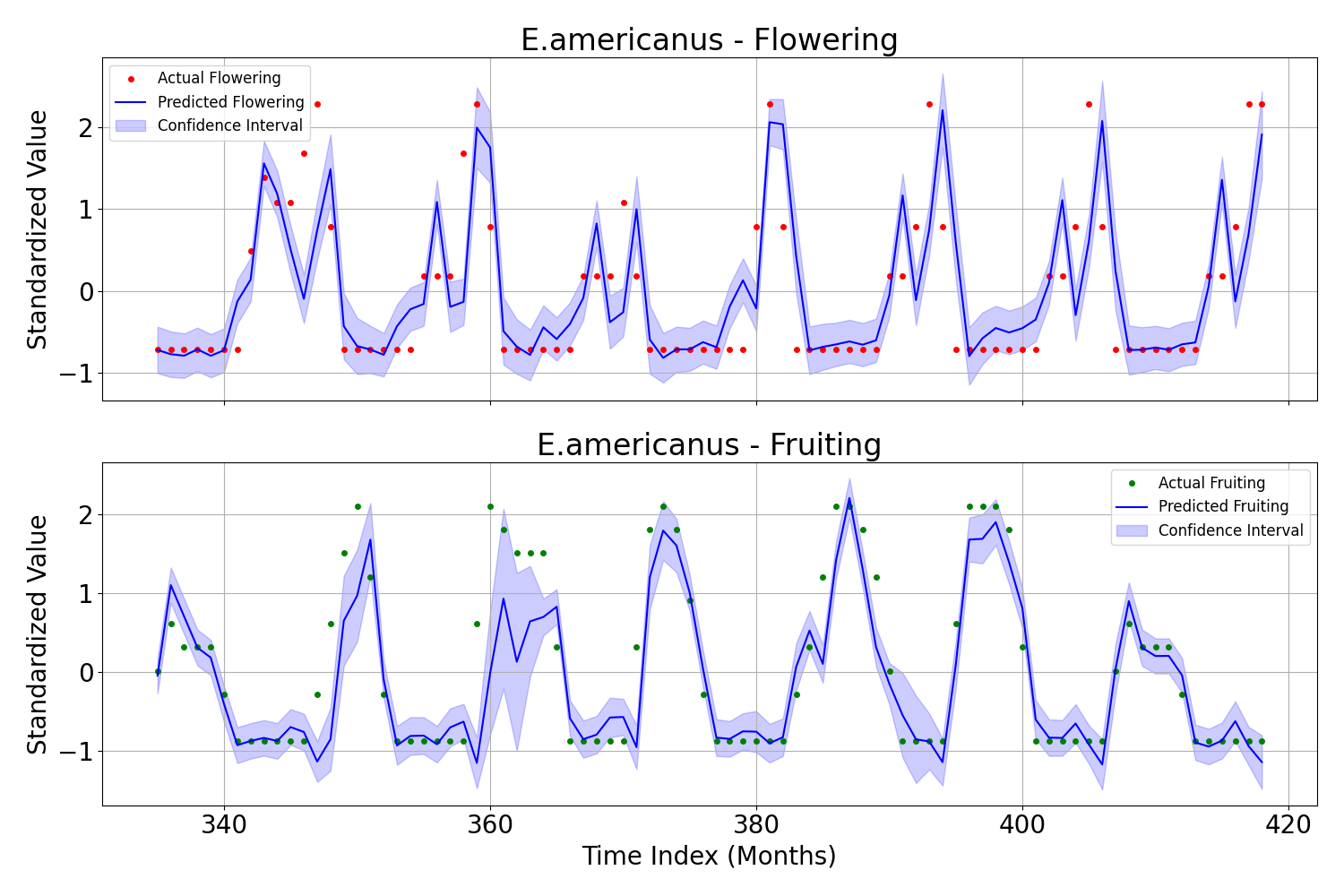}
    \end{subfigure}\hfill
    \begin{subfigure}{0.31\linewidth}
        \includegraphics[width=\textwidth]{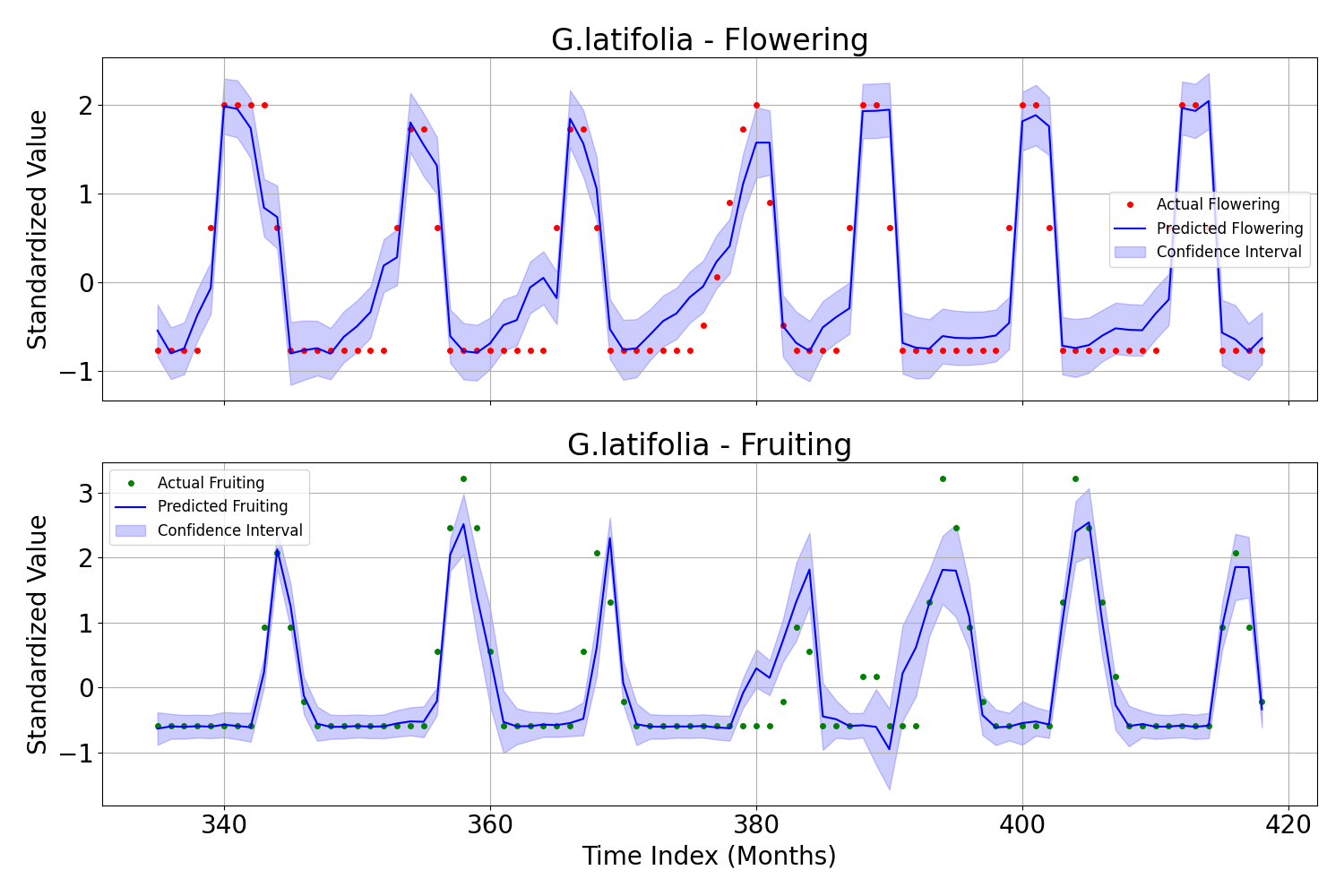}
    \end{subfigure}

    \vspace{0.2cm}

    % FILA 3
    \begin{subfigure}{0.31\linewidth}
        \includegraphics[width=\textwidth]{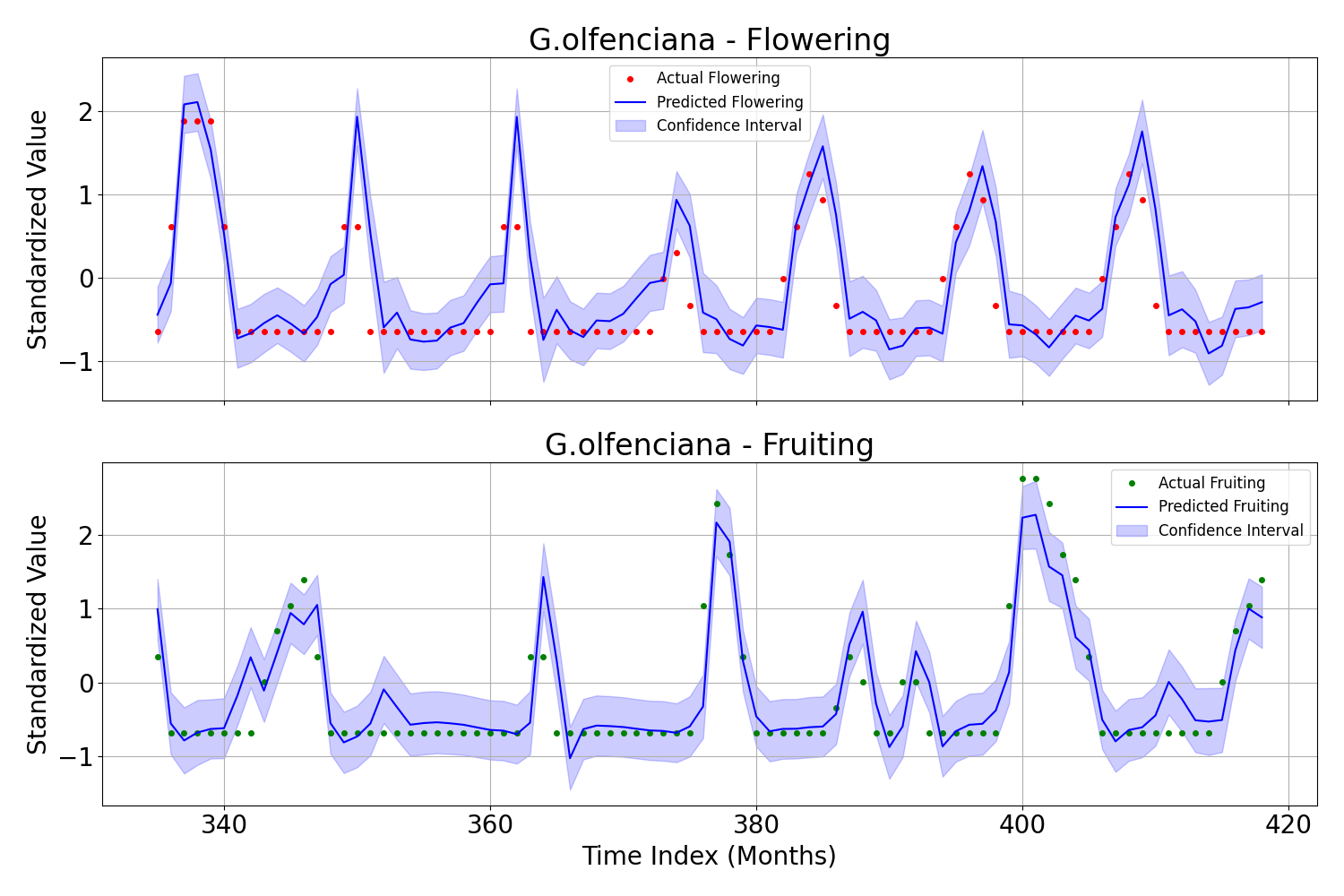}
    \end{subfigure}\hfill
    \begin{subfigure}{0.31\linewidth}
        \includegraphics[width=\textwidth]{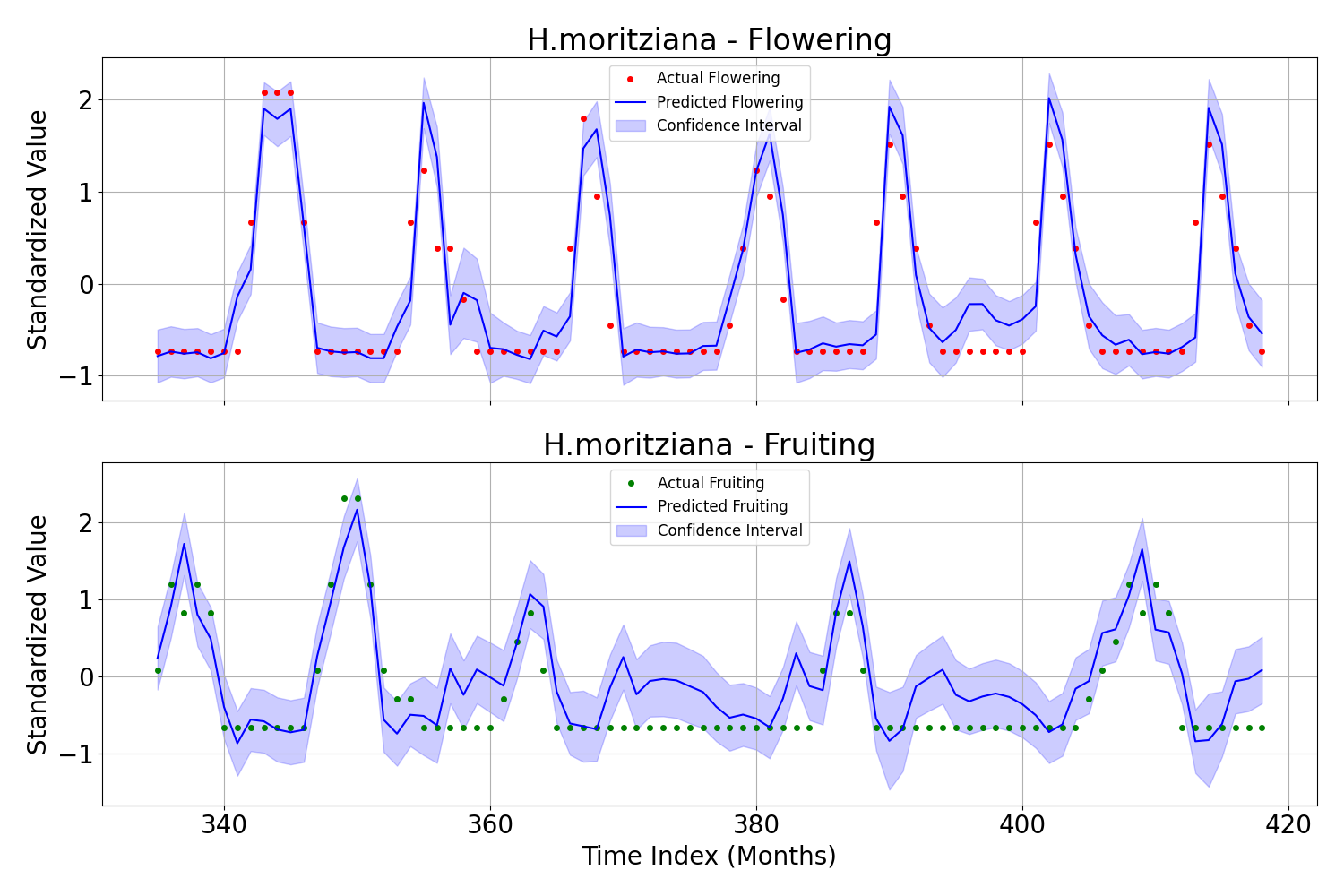}
    \end{subfigure}\hfill
    \begin{subfigure}{0.31\linewidth}
        \includegraphics[width=\textwidth]{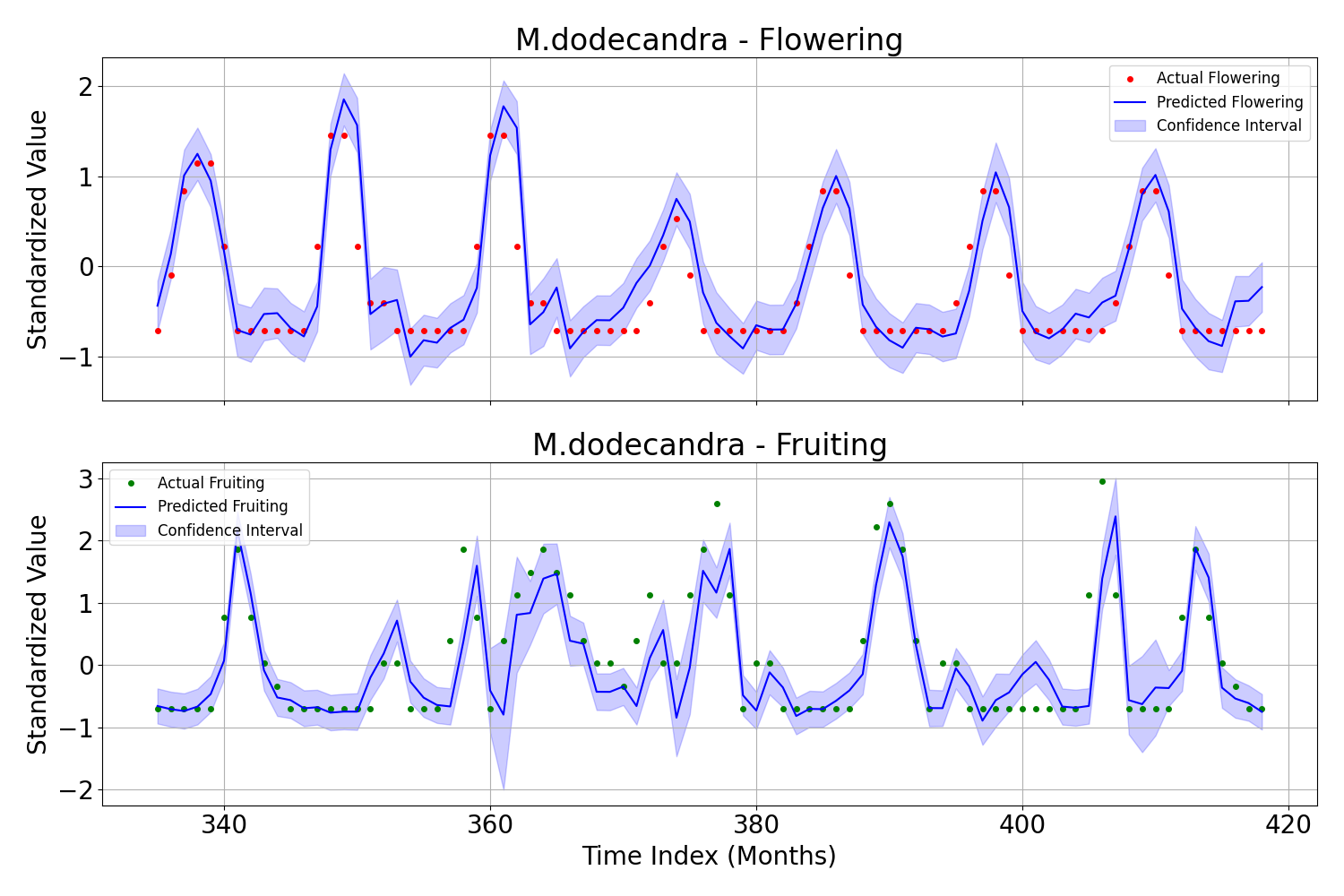}
    \end{subfigure}

    \vspace{0.2cm}

   % FILA 4
    \begin{subfigure}{0.31\linewidth}
        \includegraphics[width=\textwidth]{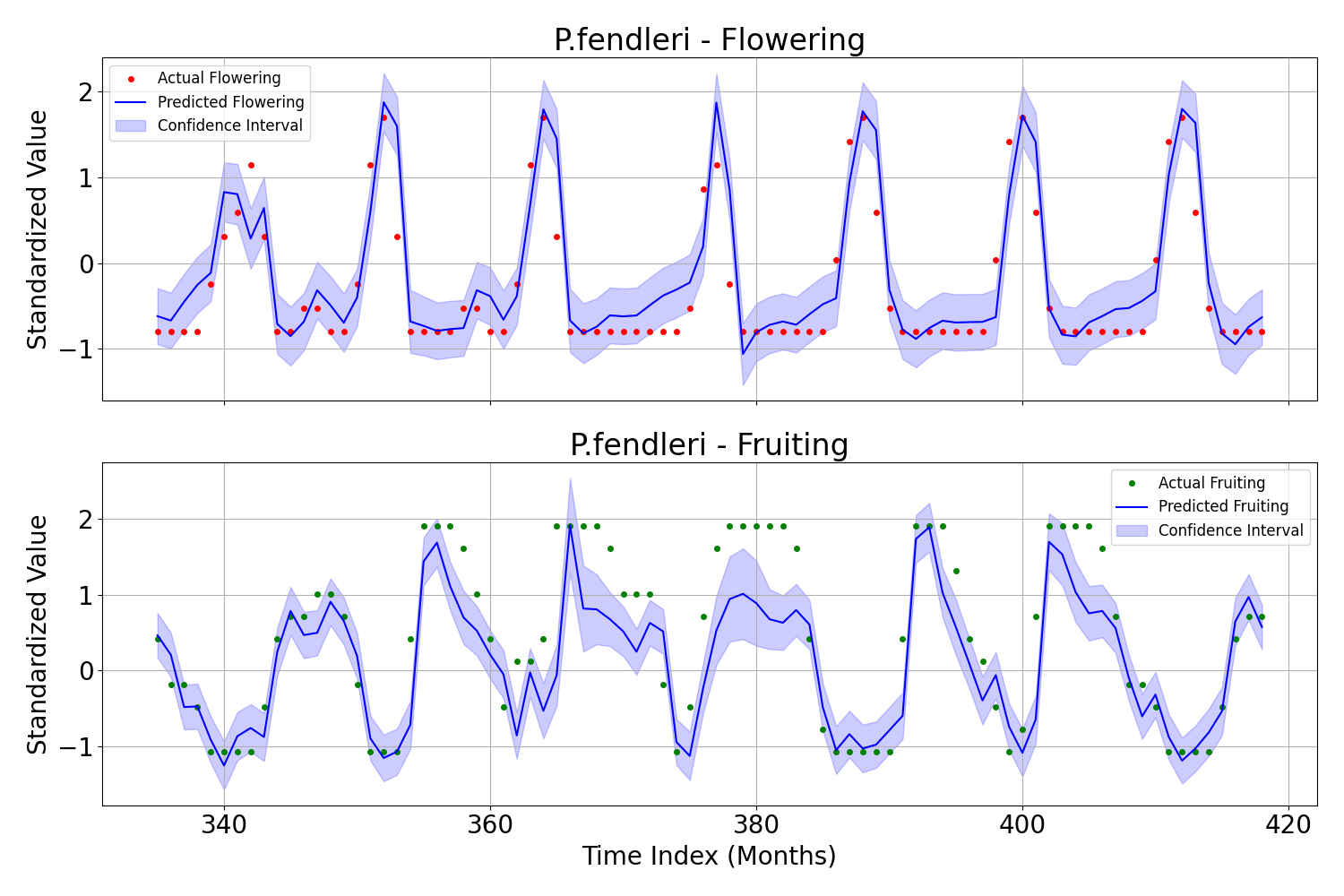}
    \end{subfigure}\hfill
    \begin{subfigure}{0.31\linewidth}
        \includegraphics[width=\textwidth]{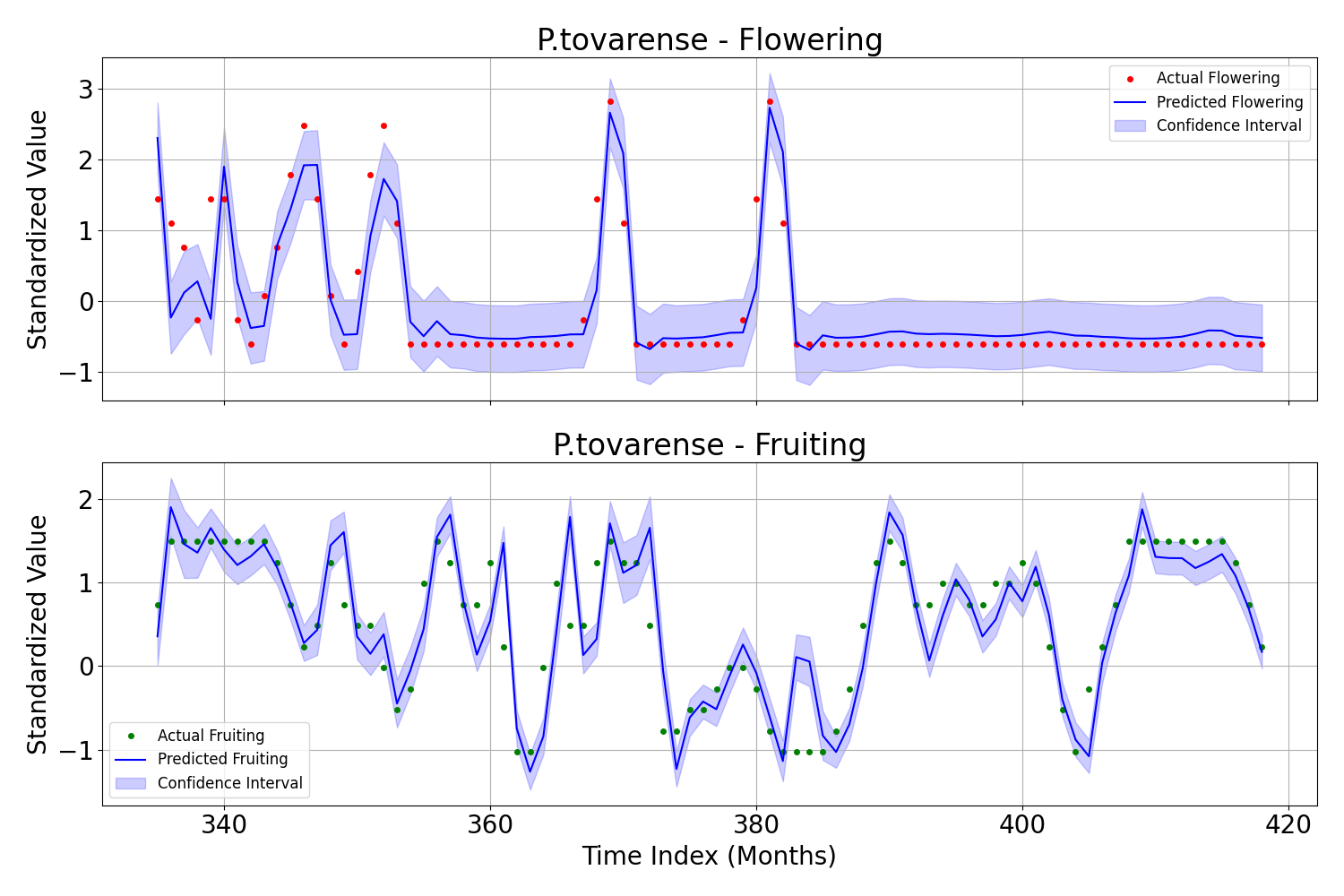}
    \end{subfigure}\hfill
    \begin{subfigure}{0.31\linewidth}
        \includegraphics[width=\textwidth]{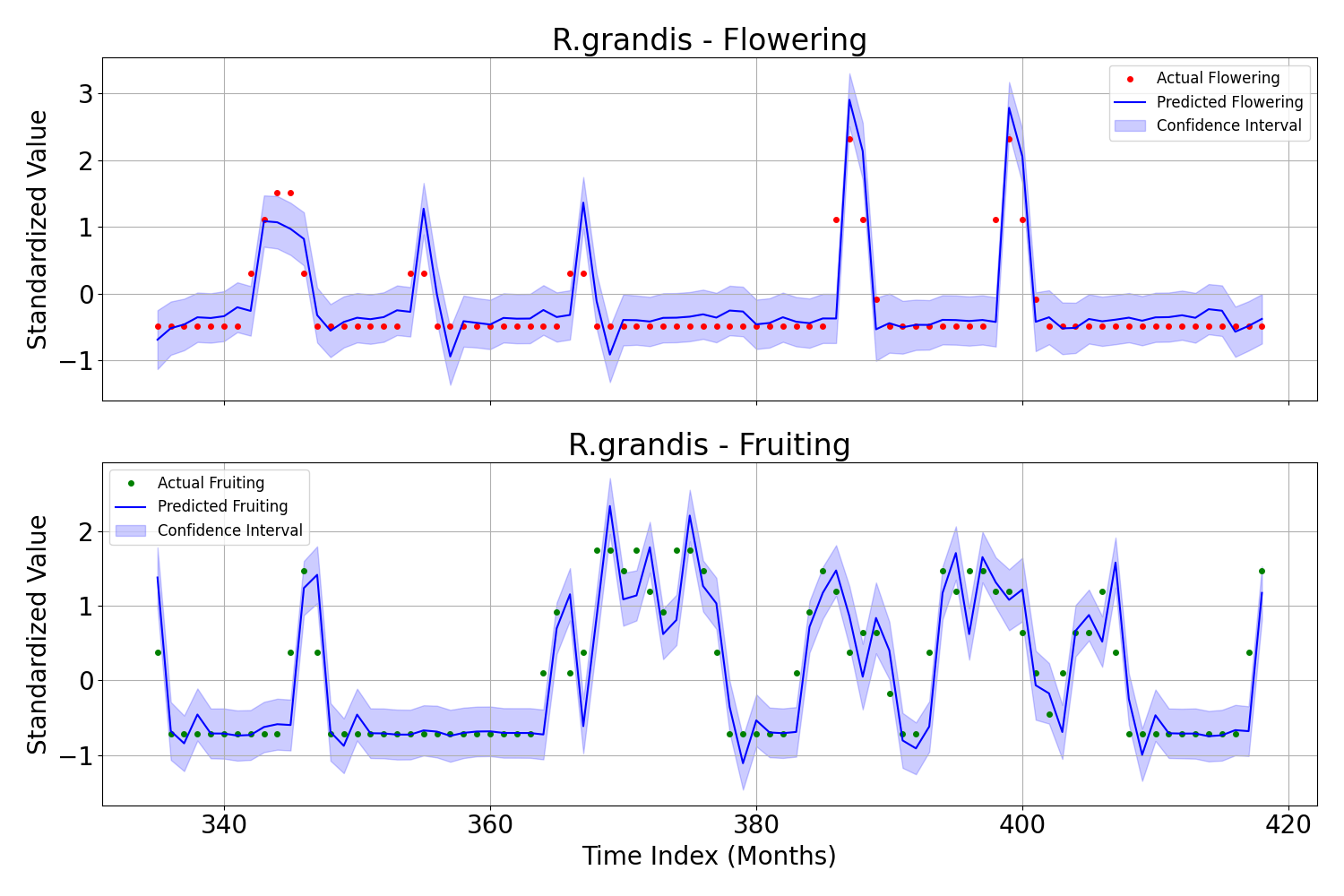}
    \end{subfigure}

    \vspace{0.2cm}

    % FILA 5 (Completada con las figuras restantes)
    \begin{subfigure}{0.31\linewidth}
        \includegraphics[width=\textwidth]{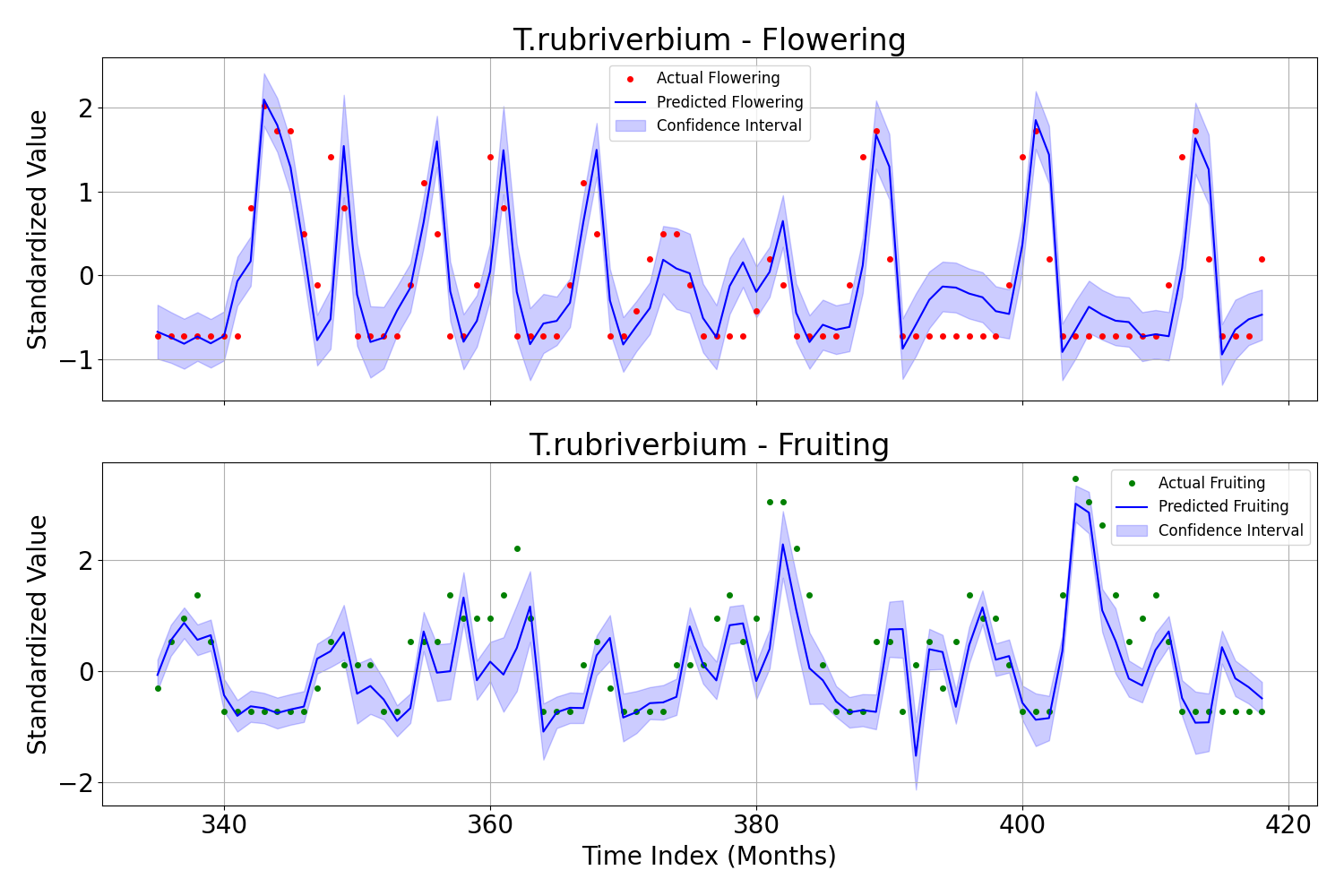}
    \end{subfigure}\hfill
    \begin{subfigure}{0.31\linewidth}
        \includegraphics[width=\textwidth]{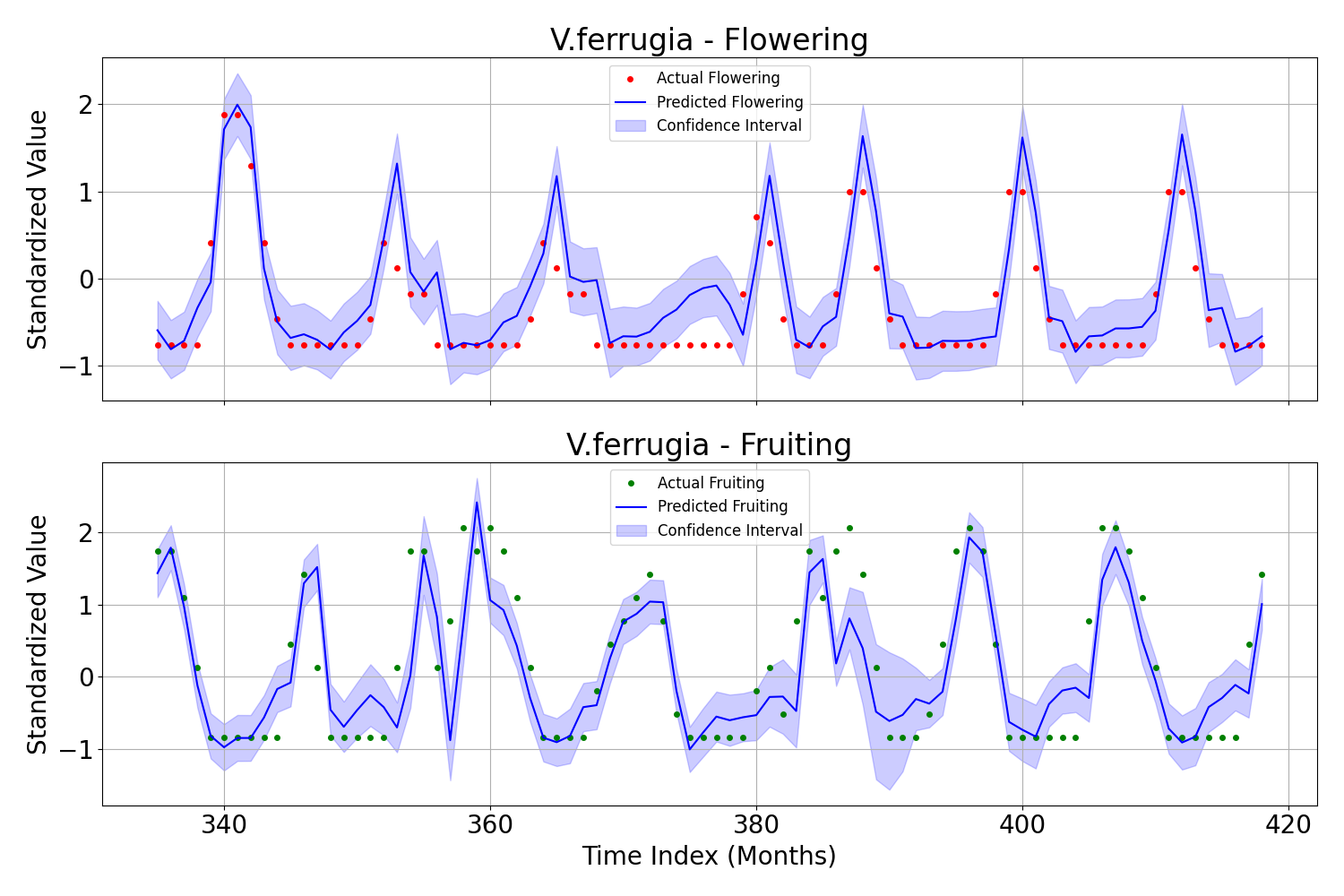} % Asegúrate de que este archivo exista
    \end{subfigure}\hfill
    \begin{subfigure}{0.31\linewidth}
        \includegraphics[width=\textwidth]{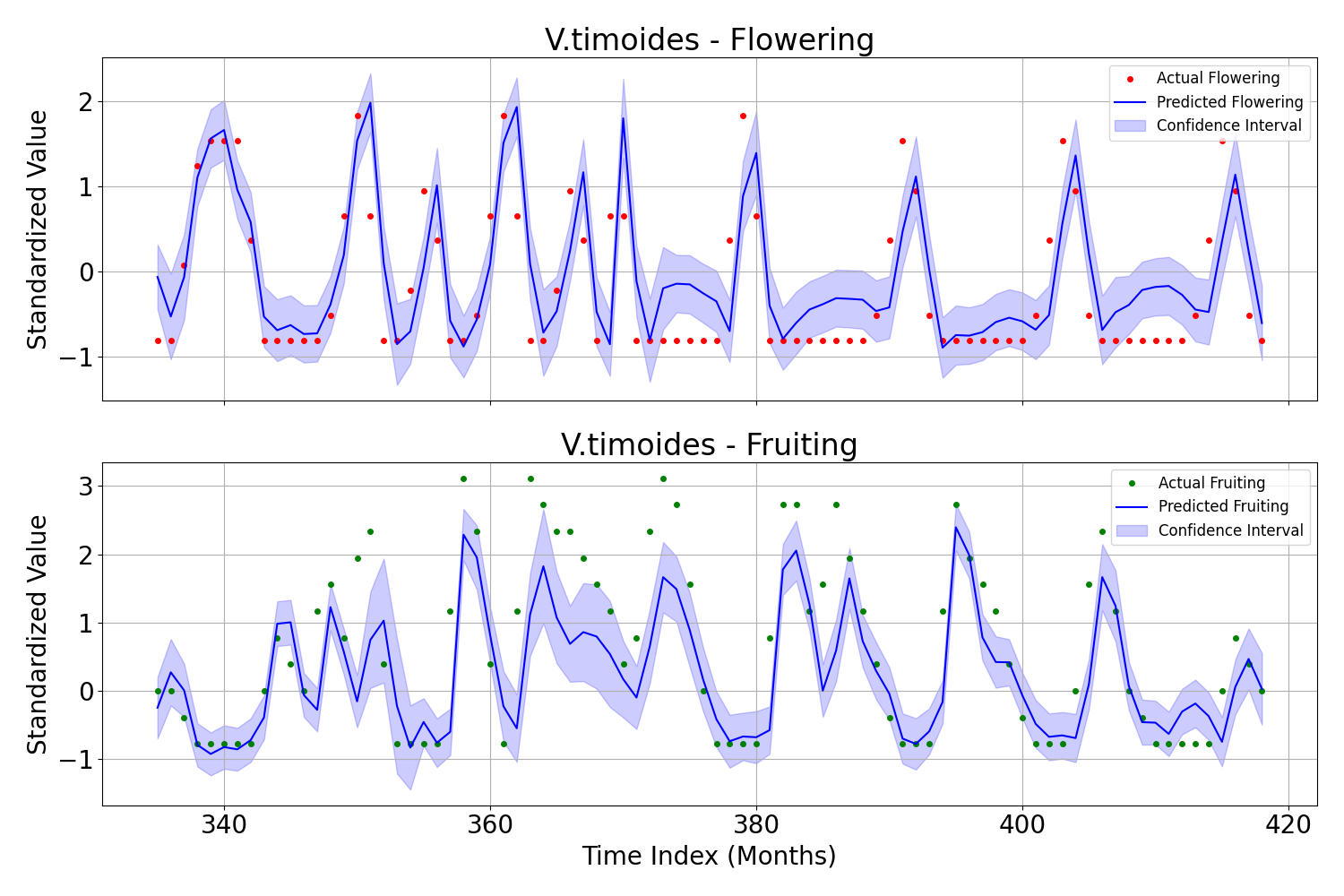} % Asegúrate de que este archivo exista
    \end{subfigure}

   \caption{GPR flowering and fruiting prediction for each specie vs. actual data with confidence interval.}
    \label{Forcast_Matrix_15}
\end{figure}

Figure \ref{Forcast_Matrix_15} shows the predictions and observations for the case of flowering and fruiting for each of the species.

To assess the temporal stability of phenological events, the Root Mean  Square Error (RMSE)  was defined as the primary metric (Table \ref{gpr_performance}), where low values indicate a high degree of fidelity to the multi-year average pattern.

\begin{table}[ht!]
\center
\begin{tabular}{lcc}
\toprule
Species & Flowering  & Fruiting \\
\midrule
A.fendleri & 0.444 &  0.581  \\
B.reticulata & 0.503  & 0.495  \\
C.tremadena & 0.444  & 0.478  \\
E.amazonicum & 0.455  & 0.637  \\
E.americanus & 0.705  & 0.546  \\
G.latifolia & 0.449  & 0.482  \\
G.olfenciana & 0.451  & 0.424  \\
H.moritziana & 0.423  & 0.441 \\
M.dodecandra & 0.340  & 0.589  \\
P.fendleri & 0.403  & 0.626  \\
P.tovarense & 0.439  & 0.448  \\
R.grandis & 0.394  & 0.470  \\
T.rubriverbium & 0.560  & 0.762 \\
V.ferrugia & 0.390  & 0.605 \\
V.timoides & 0.586  & 0.819  \\
\bottomrule
\end{tabular}
\caption{Root-mean-square error of GPR by species.}
\label{gpr_performance}
\end{table}

\begin{figure}[h]
	\begin{center}
		\includegraphics[width=14cm,height=12cm]{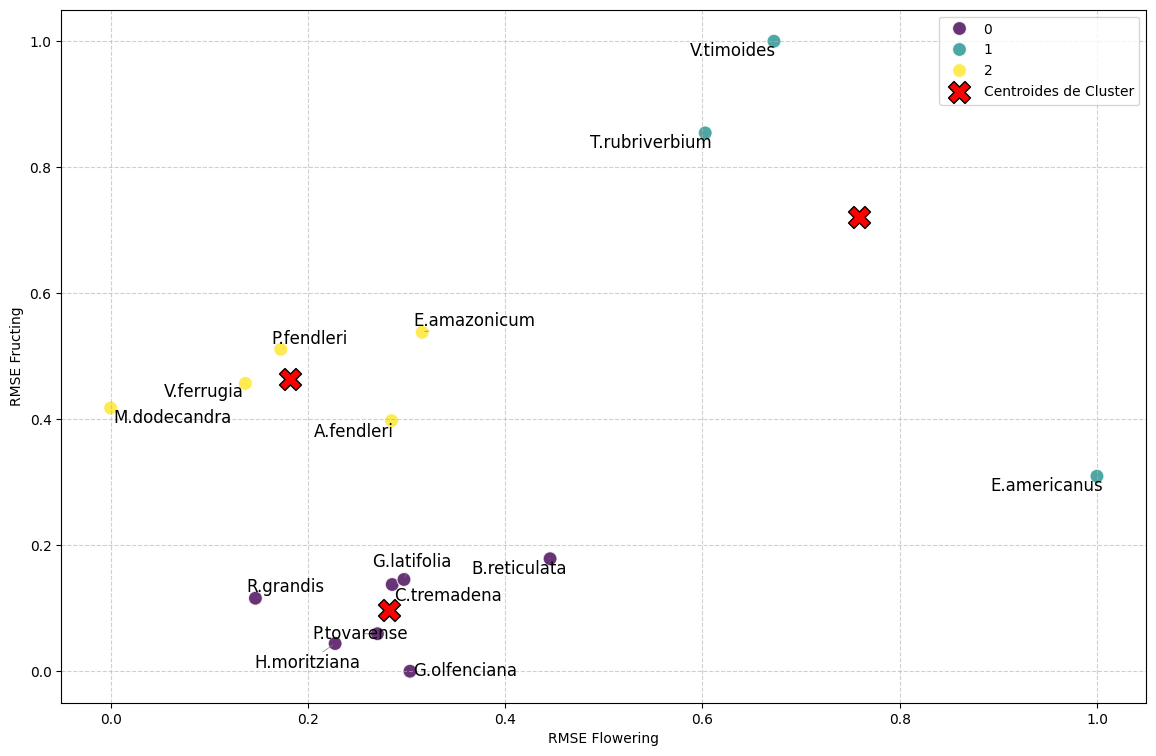}
		\caption{K-Means Clustering of species based on RMSE.}
		\label{fig:clusters}
	\end{center}
\end{figure}

In order to illustrate the performance of the GPR in a graphical way,  K-Means clustering algorithm was applied to the RMSE metrics for the flowering and fruiting stages. This identifies three distinct functional groups (Figure \ref{fig:clusters}). To avoid biases caused by differences in the scalar magnitudes of both variables, the data were normalized to the range $[0, 1]$. This normalized matrix served as input for the K-Means algorithm, allowing the phase space to be partitioned into three regions of distinct biological behavior.

This approach allows us to move beyond traditional taxonomic classification to focus on the stability and variability of the biological rhythms of the studied species.

\begin{itemize}
    \item \textbf{Cluster $C_1$ (Phenological Stability):} Composed of species such as \textit{H. moritziana}, \textit{P. tovarense}, \textit{G. latifolia}, and \textit{R. grandis}. This group is characterized by having the lowest RMSE values in both dimensions ($RMSE < 0.2$ rescaled). From an ecological perspective, this suggests high predictability in their reproductive cycles, a behavior typical of climax species in cloud forests where seasonality is less erratic.
    
    \item \textbf{Cluster $C_2$ (High Variability):} Includes \textit{V. timoides}, \textit{T. rubriverbium}, and \textit{E. americanus}. These species show significant dispersion relative to the centroids of the other groups, with marked variability in flowering. The peripheral position of \textit{E. americanus} indicates atypical behavior that could be linked to opportunistic phenological responses to interannual climate fluctuations.

    \item \textbf{Cluster $C_3$ (Intermediate Variability):} This cluster includes species such as \textit{M. dodecandra}, \textit{V. ferrugia}, and \textit{A. fendleri}. These species exhibit moderate stability in flowering but greater variation in fruiting ($0.4 < RMSE < 0.6$). The convergence of pioneer species in this group is notable, suggesting shared reproductive strategies oriented toward the dispersal of fleshy fruits in areas of secondary succession.
\end{itemize}

This type of analysis reveals significant functional convergence. Morphologically distinct species, such as the emergent tree \textit{R. grandis} and the shrub \textit{P. tovarense}, are grouped in Cluster $C_1$, indicating that they share similar temporal niches. Conversely, the segregation of species within the same genus, such as \textit{P. fendleri} (Cluster $C_3$) and a \textit{Palicourea} (Cluster $C_1$), demonstrates that phylogeny is not the sole determinant of phenological response to environmental variables in this setting.

The erratic behavior observed in certain species explains the loss of predictability in the GPR models and the increase in entropy in Cluster $C_2$. Analysis of the results suggests a phenomenon of functional decoupling in the forest, where the historical pattern of biological synchrony—the ecosystem’s “Swiss clock”—has been replaced by individual survival strategies under stress. For example, the dormancy observed in Protium and the two-year reproductive delay in Heliocarpus indicate a shift in resource allocation: species are prioritizing vegetative maintenance over mass reproduction to mitigate the metabolic cost of water deficit.

In this context, we believe that statistical error should not be interpreted merely as a model failure, but rather as a metric of water stress that disrupts the forest’s cohesive response to climate change. The algorithms’ ability to detect these changes allows us to identify how environmental pressures are forcing species to abandon ancestral rhythms in favor of plastic responses aimed at individual survival.

\section{Final comments}
Phenological dynamics of tropical cloud forests provide valuable information on the interactions that occur in the ecosystem and allow us to understand what will happen in the future with the flowering and fruiting of these forests.

By providing more accurate predictions of the timing of phenological phenomena, machine learning can help researchers and resource managers better understand how these ecosystems respond to climate change and other environmental stresses, and develop strategies to conserve and manage them in the future.

The incorporation of machine learning strategies in the analysis of data from  in tropical cloud forests, can help us to understand the complex relationships between environmental factors and phenological dynamics by identifying patterns and trends in data sets.  Particularly, these algorithms can be used to detect changes in phenological phenomena such as flowering, fruiting or leaf fall and to relate these changes to specific environmental factors such as temperature and precipitation.

In summary, machine learning has the potential to improve our understanding of the phenological dynamics of tropical cloud forests and to provide valuable information on the interactions between environmental factors and ecosystem processes. As more data become available, machine learning algorithms continue to improve, and we may have even more accurate and reliable phenological forecasts in the future.

\section*{Declaration of Interests}

\subsection*{Author Contributions}
\textbf{Conceptualization}: PG and SF, \textbf{Investigation}: PG and SF, \textbf{Methodology}: PG, \textbf{Data curation}: SF, \textbf{Writing – original draft}: PG and SF, \textbf{Writing – review and editing}: PG.

\subsection*{Competing Interests}
The authors declare that there are no conflicts of interest with respect to the publication of this article.\\

\subsection*{Declaration of generative AI use}
While preparing this paper, the authors used Gemini to edit and proofread the article. The authors have reviewed and edited the content and assume full responsibility for its accuracy and completeness.

\bibliographystyle{plainnat} % We choose the "plain" reference style
\bibliography{References} % Entries are in the refs.bib file

@article{alberton,
  author={Alberton, Bruna and Martin, Thomas CM and Da Rocha, Humberto R and Richardson, Andrew D and De Moura, Magna SB and Torres, Ricardo S and Morellato, Leonor Patricia Cerdeira},
  title={Relationship between tropical leaf phenology and ecosystem productivity using phenocameras},
  journal={Frontiers in Environmental Science},
  volume={11},
  pages={1223219},
  year={2023}
}

@article{tscharntke,
  author    = {Tscharntke, Teja},
  title     = {Disrupting plant-pollinator systems endangers food security},
  journal   = {One Earth},
  volume    = {4},
  number    = {9},
  pages     = {1217--1219},
  year      = {2021},
  month     = {sep},
  publisher = {Elsevier}
}

@article{augustin,
  author    = {Augustin, J. and Gilge, S. and Appel, H. and Dauert, U. and Endler, C. and Heesen, R. and H{\"o}flich, C. and Kuttler, W. and Schl{\"u}nzen, K. H. and Straff, W. and Werchan, B. and Werchan, M. and Zuberbier, T. and Traidl-Hoffmann, C.},
  title     = {Climate Change, Air Quality, and Pollen Allergies—State of the Art and Recommendations for Research and Public Health},
  journal   = {Allergy},
  year      = {2026},
  volume    = {81},
  number    = {3},
  pages     = {663--683}
}

@article{Bartholomew2026,
  author    = {Bartholomew, D. C. and Bittencourt, P. R. L. and Galiano Cabrera, D. and Sacatuma Cruz, R. and Chambi Paucar, J. R. and Corrales Alvarez, D. and Cosio, E. and Espinoza Otazu, B. and Mamani, D. M. and Meir, P. and Mu{\~n}oz Hermoza, G. A. and Oliveira, R. S. and Puma Vilca, L. L. and Rosalai, A. and Salas Yupayccana, C. and Salinas, N. and Sanchez Tintaya, J. and Yuca Palomino, J. A. and Metcalfe, D. B.},
  title     = {Tropical Montane Cloud Forests Have High Resilience to Five Years of Severe Soil Drought},
  journal   = {Global Change Biology},
  volume    = {32},
  number    = {1},
  pages     = {e70670},
  year      = {2026},
  month     = {Jan},
  doi       = {10.1111/gcb.70670},
  note      = {PMID: 41499343; PMCID: PMC12779095}
}

@article{flores2023extreme,
  title={Extreme drought disrupts plant phenology: Insights from 35 years of cloud forest data in Venezuela},
  author={Flores, Sa{\'u}l and Forister, Matthew L and Sulbaran, Hendrik and D{\'\i}az, Rodrigo and Dyer, Lee A},
  journal={Ecology},
  volume={104},
  number={5},
  pages={e4012},
  year={2023},
  publisher={Wiley Online Library}
}

@article{Garcia-1996,
  title = {Local Optimal Metrics and Nonlinear Modeling of Chaotic Time Series},
  author = {Garc\'{\i}a, P. and Jim\'enez, J. and Marcano, A. and Moleiro, F.},
  journal = {Phys. Rev. Lett.},
  volume = {76},
  issue = {9},
  pages = {1449--1452},
  numpages = {0},
  year = {1996},
  month = {Feb},
  publisher = {American Physical Society}
}

@article{williams1999,
  title={Leaf dynamics in a tropical cloud forest: phenology, herbivory, and life span},
  author={Williams-Linera, Guadelupe},
  journal={Selbyana},
  pages={98--105},
  year={1999},
  publisher={JSTOR}
}

@article{sullivan2023cascading,
  title={Cascading effects of climate change: new advances in drivers and shifts of tropical reproductive phenology},
  author={Sullivan, Megan K and Fayolle, Adeline and Bush, Emma and Ofosu-Bamfo, Bismark and Vleminckx, Jason and Metz, Margaret R and Queenborough, Simon A},
  journal={Plant Ecology},
  pages={1--13},
  year={2023},
  publisher={Springer}
}

@article{bucher2021timing,
  title={The timing of leaf senescence relates to flowering phenology and functional traits in 17 herbaceous species along elevational gradients},
  author={Bucher, Solveig Franziska and R{\"o}mermann, Christine},
  journal={Journal of Ecology},
  volume={109},
  number={3},
  pages={1537--1548},
  year={2021},
  publisher={Wiley Online Library}
}

@article{fu2022plant,
  title={Plant phenology shifts and their ecological and climatic consequences},
  author={Fu, Yongshuo H and Prev{\'e}y, Janet S and Vitasse, Yann},
  journal={Frontiers in Plant Science},
  volume={13},
  pages={1071266},
  year={2022},
  publisher={Frontiers}
}

@article{davis2022new,
  title={New directions in tropical phenology},
  author={Davis, Charles C and Lyra, Goia M and Park, Daniel S and Asprino, Renata and Maruyama, Rog{\'e}rio and Torquato, D{\'e}bora and Cook, Benjamin I and Ellison, Aaron M},
  journal={Trends in ecology \& evolution},
  year={2022},
  publisher={Elsevier}
}

@article{auger2021guide,
  title={A guide to state--space modeling of ecological time series},
  author={Auger-M{\'e}th{\'e}, Marie and Newman, Ken and Cole, Diana and Empacher, Fanny and Gryba, Rowenna and King, Aaron A and Leos-Barajas, Vianey and Mills Flemming, Joanna and Nielsen, Anders and Petris, Giovanni and others},
  journal={Ecological Monographs},
  volume={91},
  number={4},
  pages={e01470},
  year={2021},
  publisher={Wiley Online Library}
}

@article{li2023machine,
  title={A machine learning approach targeting parameter estimation for plant functional type coexistence modeling using ELM-FATES (v2. 0)},
  author={Li, Lingcheng and Fang, Yilin and Zheng, Zhonghua and Shi, Mingjie and Longo, Marcos and Koven, Charles D and Holm, Jennifer A and Fisher, Rosie A and McDowell, Nate G and Chambers, Jeffrey and others},
  journal={Geoscientific Model Development},
  volume={16},
  number={14},
  pages={4017--4040},
  year={2023},
  publisher={Copernicus Publications G{\"o}ttingen, Germany}
}

@article{weinan2017proposal,
  title={A proposal on machine learning via dynamical systems},
  author={Weinan, Ee},
  journal={Communications in Mathematics and Statistics},
  volume={1},
  number={5},
  pages={1--11},
  year={2017}
}

@article{pourmohammad2023dynamical,
  title={A dynamical systems approach to machine learning},
  author={Pourmohammad Azizi, S and Neisy, Abdolsadeh and Ahmad Waloo, Sajad},
  journal={International journal of computational methods},
  volume={20},
  number={09},
  pages={2350007},
  year={2023},
  publisher={World Scientific}
}

@article{guan2023study,
  title={A study of forest phenology prediction based on GRU models},
  author={Guan, Peng and Zhu, Lichen and Zheng, Yili},
  journal={Applied Sciences},
  volume={13},
  number={8},
  pages={4898},
  year={2023},
  publisher={MDPI}
}

@INPROCEEDINGS{Wang2019,
  author={Wang, Yuchao and Liu, Shuhe and Sun, Yunpeng and Fu, Huixuan},
  booktitle={2019 IEEE International Conference on Mechatronics and Automation (ICMA)}, 
  title={Phenological Prediction Algorithm Based on Deep Learning}, 
  year={2019},
  volume={},
  number={},
  pages={589-593}
}

@article{verrelst2013gaussian,
  title={Gaussian processes uncertainty estimates in experimental Sentinel-2 LAI and leaf chlorophyll content retrieval},
  author={Verrelst, Jochem and Rivera, Juan Pablo and Moreno, Jos{\'e} and Camps-Valls, Gustavo},
  journal={ISPRS journal of photogrammetry and remote sensing},
  volume={86},
  pages={157--167},
  year={2013},
  publisher={Elsevier}
}

@article{yates2023cross,
  title={Cross validation for model selection: a review with examples from ecology},
  author={Yates, Luke A and Aandahl, Zach and Richards, Shane A and Brook, Barry W},
  journal={Ecological Monographs},
  volume={93},
  number={1},
  pages={e1557},
  year={2023},
  publisher={Wiley Online Library}
}

@incollection{kalnay2018ncep,
  title={The NCEP/NCAR 40-year reanalysis project},
  author={Kalnay, Eugenia and Kanamitsu, Masao and Kistler, Robert and Collins, William and Deaven, Dennis and Gandin, Lev and Iredell, Mark and Saha, Suranjana and White, Glenn and Woollen, John and others},
  booktitle={Renewable energy},
  pages={Vol1\_146--Vol1\_194},
  year={2018},
  publisher={Routledge}
}

@inproceedings{takens2006detecting,
  title={Detecting strange attractors in turbulence},
  author={Takens, Floris},
  booktitle={Dynamical Systems and Turbulence, Warwick 1980: proceedings of a symposium held at the University of Warwick 1979/80},
  pages={366--381},
  year={2006},
  organization={Springer}
}

@article{Garcia-2022,
title = {Modeling systems with machine learning based differential equations},
journal = {Chaos, Solitons \& Fractals},
volume = {165},
pages = {112872},
year = {2022},
issn = {0960-0779},
author = {P. García}
}

@Inbook{Williams1998,
author="Williams, C. K. I.",
editor="Jordan, Michael I.",
title="Prediction with Gaussian Processes: From Linear Regression to Linear Prediction and Beyond",
bookTitle="Learning in Graphical Models",
year="1998",
publisher="Springer Netherlands",
address="Dordrecht",
pages="599--621",
isbn="978-94-011-5014-9"
}

@book{Ramusen,
  title     = "Gaussian Processes for Machine Learning",
  author    = " Rasmussen, Carl Edward and Williams, Christopher K. I. ",
  year      =  2005,
  publisher = "MIT Press",
  address   = "Cambridge"
}

@book{cristianini2000introduction,
  title={An Introduction to Support Vector Machines and Other Kernel-based Learning Methods},
  author={Cristianini, N. and Shawe-Taylor, J.},
  isbn={9780521780193},
  lccn={99054716},
  year={2000},
  publisher={Cambridge University Press}
}

@article{Zalika,
title = {Modelling of weather variability effect on fitophenology},
author = {Zalika Črepinšek and Lučka Kajfež-Bogataj and Klemen Bergant},
journal = {Ecological Modelling},
volume = {194},
number = {1},
pages = {256-265},
year = {2006},
note = {Special Issue on the Fourth European Conference on Ecological Modelling},
issn = {0304-3800}
}

@article{Garcia-2007,
  author    = {Garc{\'\i}a, P. and Merlitti, A.},
  title     = {Haar basis and nonlinear modeling of complex systems},
  journal   = {The European Physical Journal Special Topics},
  volume    = {143},
  number    = {1},
  pages     = {261--264},
  year      = {2007},
  publisher = {Springer Science and Business Media LLC}
}

@article{kimeldorf1971some,
  title={Some results on Tchebycheffian spline functions},
  author={Kimeldorf, George S. and Wahba, Grace},
  journal={Journal of Mathematical Analysis and Applications},
  volume={33},
  number={1},
  pages={82--95},
  year={1971},
  publisher={Elsevier},
  doi={10.1016/0022-247X(71)90184-3}
}

@inproceedings{scholkopf2001generalized,
  title={A Generalized Representer Theorem},
  author={Sch{\"o}lkopf, Bernhard and Herbrich, Ralf and Smola, Alex J.},
  booktitle={Computational Learning Theory (COLT / EML)},
  series={Lecture Notes in Computer Science},
  volume={2011},
  pages={416--426},
  year={2001},
  publisher={Springer},
  doi={10.1007/3-540-44581-1_27}
}
\end{document}